\documentclass[10pt,twocolumn,letterpaper]{article}

\usepackage{iccv}
\usepackage{times}
\usepackage{epsfig}
\usepackage{graphicx}
\usepackage{amsmath}
\usepackage{amssymb}
\usepackage{multirow}
\usepackage{enumitem}

\usepackage[pagebackref=true,breaklinks=true,letterpaper=true,colorlinks,bookmarks=false]{hyperref}

\iccvfinalcopy 

\def\iccvPaperID{4957} 

\ificcvfinal\fi

\begin{document}

\title{Diverse Inpainting and Editing with GAN Inversion}

\author{Ahmet Burak Yildirim$^{*}$ \qquad Hamza Pehlivan\thanks{Joint first authors, contributed equally.}  \qquad
Bahri Batuhan Bilecen  \qquad
Aysegul Dundar \\
Bilkent University\\
\tt\small \{a.yildirim, hamza.pehlivan, batuhan.bilecen\}@bilkent.edu.tr\\
\tt\small adundar@cs.bilkent.edu.tr\\}

\maketitle

\begin{abstract}

Recent inversion methods have shown that real images can be inverted into StyleGAN's latent space and numerous edits can be achieved on those images thanks to the semantically rich feature representations of well-trained GAN models. 
However, extensive research has also shown that image inversion is challenging due to the trade-off between high-fidelity reconstruction and editability.
In this paper, we tackle an even more difficult task, inverting erased images into GAN's latent space for realistic inpaintings and editings. 
Furthermore, by augmenting inverted latent codes with different latent samples, we achieve diverse inpaintings.
Specifically, we propose to learn an encoder and mixing network to combine encoded features from erased images with StyleGAN's mapped features from random samples. 
To encourage the mixing network to utilize both inputs, we train the networks with generated data via a novel set-up.
We also utilize higher-rate features to prevent color inconsistencies between the inpainted and unerased parts.
 We run extensive experiments and compare our method with state-of-the-art inversion and inpainting methods. Qualitative metrics and visual comparisons show significant improvements.

\end{abstract}

\section{Introduction}

Generative Adversarial Networks (GANs) achieve impressive results on unconditional ~\cite{ karras2019style, zhang2019self, karras2020analyzing,yu2021dual} and conditional image generations \cite{wang2018high, dundar2020panoptic, liu2022partial}, inpainting \cite{yu2019free, li2020recurrent,zhao2021large}, and image editing tasks \cite{abdal2021styleflow, chen2022exploring,dalva2022vecgan}. 
Traditionally, each of these tasks has been explored with a dedicated network and training pipeline.
However, recently it has been shown that high-quality image editing can be achieved with well-trained GAN models and especially by StyleGAN networks \cite{karras2019style, karras2020analyzing} that are trained to generate images without a condition \cite{tov2021designing, wang2022high}.
This approach achieves numerous edits via semantically rich feature representations of well-trained GANs \cite{shen2020interpreting, harkonen2020ganspace,patashnik2021styleclip,chen2022exploring}.
In this work, we are interested in taking this direction one step further and learning an encoder for a pretrained GAN that can achieve high-quality image inversion, diverse inpainting, and editing under one framework.

\newcommand{\interpfigt}[1]{\includegraphics[trim=0 0 0cm 0, clip, width=2.6cm]{#1}}

\begin{figure}[t]
\centering
\includegraphics[width=\linewidth]{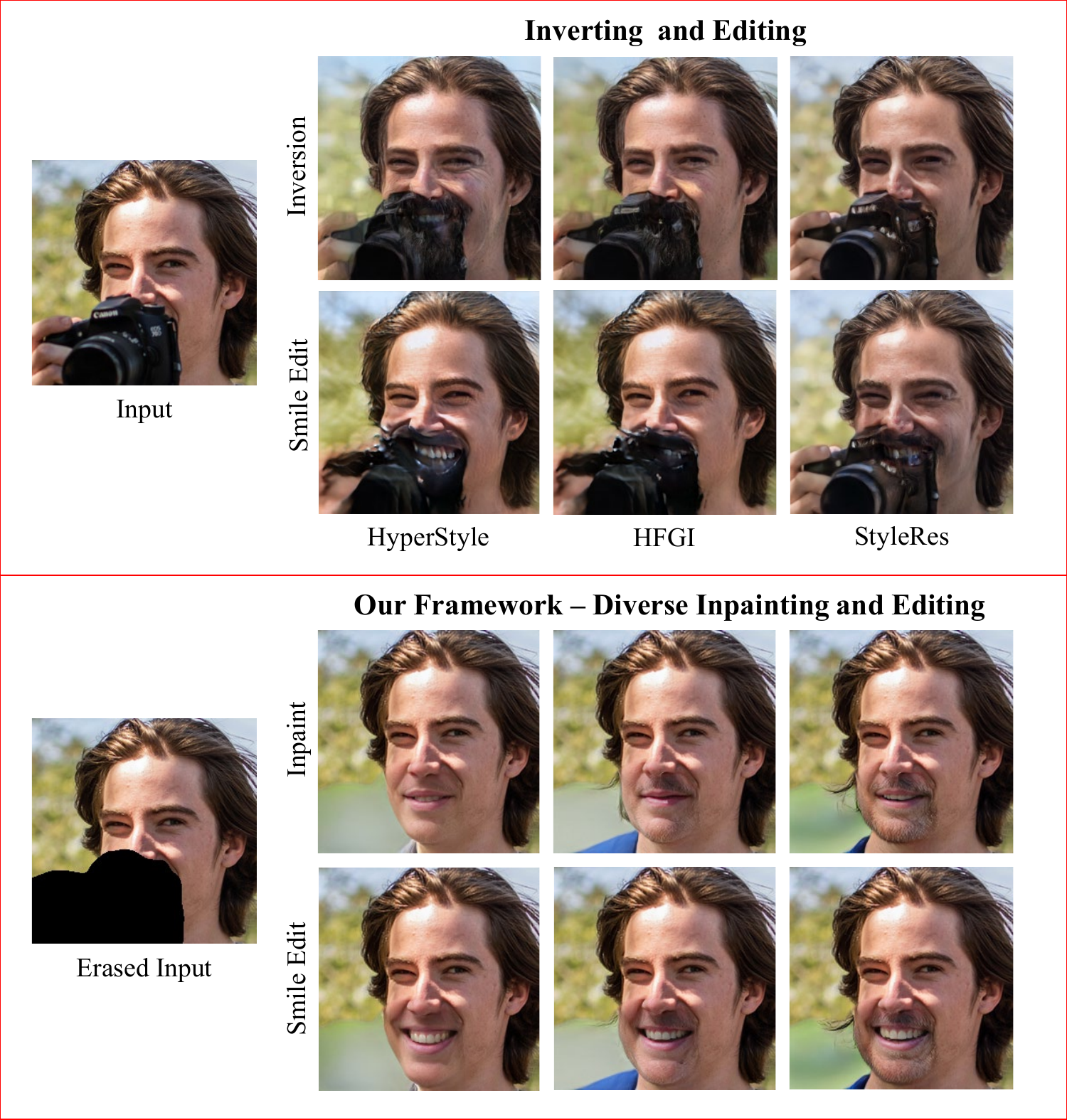}
\caption{ Our framework achieves diverse inpainting and editing with GAN inversion.
Compared to other works that are proposed to invert and edit images with StyleGAN (HyperStyle \cite{alaluf2022hyperstyle}, HFGI \cite{wang2022high}, StyleRes \cite{pehlivan2022styleres}), our framework has advantages as in the presented example.
}
\label{fig:teaser}
\end{figure}


To benefit from GAN's image editing capabilities, extensive research is conducted on image inversion algorithms to find the corresponding latent codes that will generate a particular real image \cite{creswell2018inverting, roich2022pivotal, zhu2020domain, tov2021designing}. It is observed that there exists a trade-off between image reconstruction fidelity and image editing quality \cite{tov2021designing}.
Studies show that low-rate latent spaces ($z$, $W$, or $W ^+$ for StyleGANs) are limited in their expressiveness power, and not every image can be inverted with high-fidelity reconstruction to GAN's natural low-rate latent space \cite{tov2021designing}.
When images are not projected to GAN's natural latent space, even when an image is reconstructed with high fidelity, it will not be editable.
Methods are proposed to skip spatially higher resolution features (higher-rate latent codes) from an encoder to a GAN generator for better reconstruction and editing properties \cite{wang2022high,alaluf2022hyperstyle}. 
In this work, we are interested in a more challenging scenario; learning an image encoder that can project an erased image to a well-trained GAN's natural latent space.
This framework provides new capabilities for image editing, as shown in Fig. \ref{fig:teaser}.

Recently, GAN inversion-based image inpainting methods have been proposed \cite{yu2022high, wang2022dual}. 
These methods optimize an image encoder and learn skip connections to pretrained StyleGAN model. 
Even though promising results are achieved, these methods model image inpainting in a deterministic way.
They are trained to reconstruct the original image from the erased ones without any stochasticity. 
They are not regularized to project the image into GAN's natural latent space and do not enjoy the editing capabilities of them.

In this work, we propose a framework that achieves high-quality image inversion, diverse inpainting, and editing simultaneously. 
We design an encoder architecture that takes an erased input image and encodes latent codes for the visible parts. We also design a mixing network that combines randomly sampled latent codes with encoded ones.
This setup allows the model to output diverse results.
However, we find that the framework learns to ignore the randomly sampled latent codes and has the tendency to output deterministic results.
To achieve diversity, we propose to train the framework with a novel design.
We use generated data and make use of $W^+$ and generated image pairs to regularize the network to use both inputs; erased image and sampled $W^+$.
Secondly, to achieve high fidelity inversion of unerased pixels and to prevent the color discrepancy between the unerased and erased parts, we learn higher dimension latent codes.
In summary, our main contributions are as follows:
\begin{itemize}[leftmargin=*]
    \item We propose a novel framework for image inpainting with GAN inversion.
    Our framework includes an encoder to embed images and a mixing network to combine them with randomly sampled latent codes to achieve diversity.
    The mixing network has a gating mechanism that improves the results.
    \item To achieve diversity, we propose a novel set-up to train the networks.
    We use GAN-generated images and train the network with full image reconstruction and valid pixel image reconstruction depending on if we feed the same latent code to the mixing network that is used to generate the input image or not. 
    \item We conduct extensive experiments to show the effectiveness of our framework and achieve significant improvements over state-of-the-art models for image inpainting. 
    Additionally, we show our framework can achieve diverse inpainting and editing under one framework  as shown in Fig. \ref{fig:teaser}.

\end{itemize}

\section{Related Work}

\textbf{GAN Inversion and Editing.} Recently, extensive research is conducted on image editing tasks both via end-to-end trained image translation networks \cite{starganv2,xiao2018elegant,zhang2018generative,li2021image,gao2021high,dalva2022vecgan} and latent space manipulation of pretrained GANs \cite{creswell2018inverting, roich2022pivotal, zhu2020domain, tov2021designing, wang2022high}.
Pretrained GANs, especially StyleGAN-based methods \cite{karras2019style, karras2020analyzing} are shown to organize their latent space in a semantically meaningful way with disentangled representations.
This feature enables them to achieve various edits that are beyond what annotated datasets offer \cite{voynov2020unsupervised, wu2021stylespace, patashnik2021styleclip, harkonen2020ganspace, chen2022exploring}. 
However, the challenge in using these pretrained GANs for real image editing is that one needs to invert images to the natural latent space of GANs and there is known to be a trade-off between reconstruction fidelity of input images and their editability \cite{zhu2020domain, richardson2021encoding,  tov2021designing}. 
Different encoders and training schemes are proposed to tackle the inversion problem with great improvements and the inversion problem is still an active research area \cite{wang2022high, alaluf2021restyle, alaluf2022hyperstyle, pehlivan2022styleres}. 
This work aims to solve a more challenging inversion problem with input images that have missing pixels. This task is referred to as image inpainting or image completion and to the best of our knowledge, our work is the first to explore it with pretrained StyleGAN with the high reconstruction and editability goal.

\textbf{Inpainting.} Inpainting missing pixels especially large-scale holes receives a significant amount of attention from both the research community and industry.
This task requires filling in missing pixels in a semantically meaningful and multi-modular way.
However, most of the proposed methods are deterministic and struggle with large-scale holes \cite{pathak2016context, liu2018image, yu2019free, li2020recurrent, liu2022partial}.
Recently, CoModGAN \cite{zhao2021large} is proposed with an architecture and training pipeline similar to StyleGAN but with skip connections from the encoder to the generator to pass the valid pixels to the generator without a bottleneck. 
CoModGAN achieves diverse results.
Diffusion models have also shown great success in diverse image inpainting \cite{lugmayr2022repaint, rombach2022high, yildirim2023inst}.
Nevertheless, none of these methods can achieve diverse inpainting and enable user-controlled editing of images with one network simultaneously which is the goal of our work.

\begin{figure*}[t]
\centering
\includegraphics[width=\linewidth]{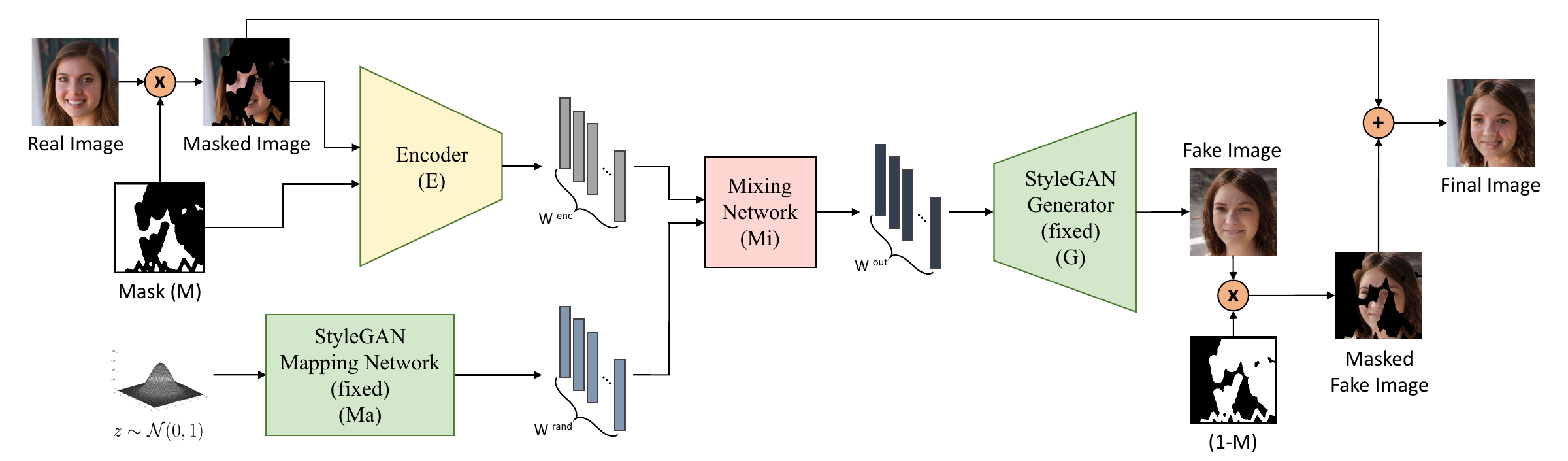}
\caption{First stage framework includes trainable image encoder and mixing network, and frozen StyleGAN's mapping and generator networks. Our encoder takes an erased image and binary mask to embed the image into StyleGAN's latent space.
We also sample a latent code via the mapping network to achieve stochasticity. 
The mixing network combines the available information of the erased image from the encoder and the missing part from the mapping network.
The mixed encoded latent representations are fed to the Generator via the instance normalization layers to output the fake image. There is a final step where the input image and the fake image are combined based on the mask.
}
\label{fig:first-stage-arch}
\end{figure*}

\section{Method}

We propose a two-stage training pipeline. In the first stage, we learn a base encoder and mixing network to achieve diverse results as presented in Section \ref{sec:first}.
In the second stage, Section \ref{sec:second}, we learn skip connections to the generator to achieve high-fidelity reconstructions and seamless transitions between unerased and erased pixels.

\subsection{First Stage: Diverse Results}
\label{sec:first}

Our method utilizes an architecture that includes an image encoder (E), mixing network (Mi), and StyleGAN2's mapping (Ma) and generator (G) networks \cite{karras2020analyzing}. We freeze the StyleGAN's mapping and generator networks as shown in Fig. \ref{fig:first-stage-arch}.
We obtain erased images by multiplying them with a binary mask $\text{M}$ that defines the valid and erased pixels; $ \text{I}^e = \text{M} \odot \text{I}$.
Before feeding the erased image to the encoder, we concatenate the binary mask $\text{M}$ with it. 
We set a simple encoder \cite{richardson2021encoding} to project our erased images.
Our encoder embeds erased images into $ \text{W}^+$ latent space.
Our goal is to achieve diversity in our inpainting results.
For that, we set a second pathway that utilizes StyleGAN's mapping network to sample random $z$'s and obtains $ \text{W}^+$ codes that lie in the natural space of StyleGAN.
The encoded, $ \text{W}^{enc} $, and mapped, $ \text{W}^{rand} $, latent codes are fed to the mixing network. 
We expect the mixing network to take the information of what is visible from the masked image and combine the missing information with the mapped latent codes.
We equip the mixing network with a neural network and gating mechanism as follows: 

\begin{equation}
\label{eq:gating}
\begin{split}
    \text{W}^{comb}, g = \text{NN}( \text{W}^{enc} , \text{W}^{rand} )\\ 
    \text{W}^{out} = \sigma(g) \cdot \text{W}^{comb} + (1-\sigma(g)) \cdot \text{W}^{rand} 
\end{split}
\end{equation}

where $\sigma(.)$ is the sigmoid function.
$\text{W}^{out}$ goes to the instance normalization layers of StyleGAN generator.

\begin{figure*}[t]
\centering
\includegraphics[width=\linewidth]{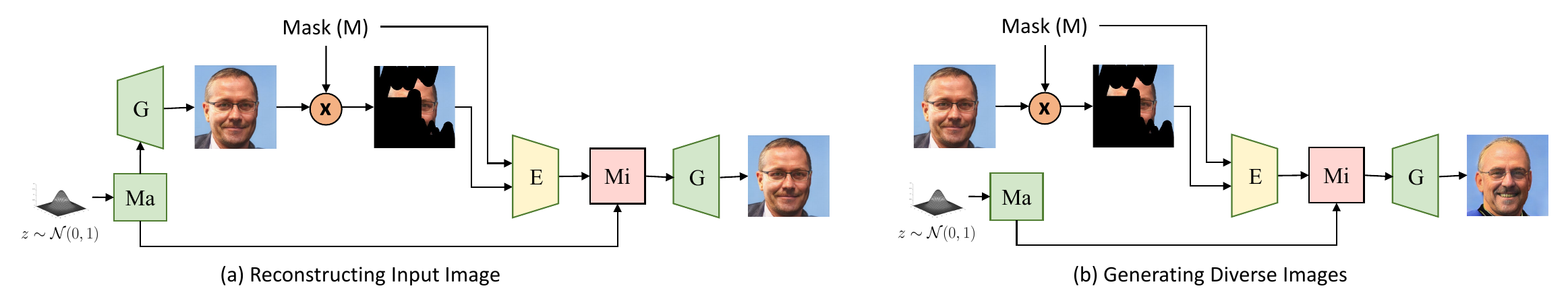}
\caption{During training, we generate images with the StyleGAN generator. 
If we feed the mixing network (Mi) with the sampled $z$ that generated the input image, we expect the output to match the input.
The reconstruction losses are applied for both valid and invalid pixels.
If we sample a new $z$ as in (b), we expect to output diverse results and only apply image reconstruction loss for valid pixels and a GAN loss for the overall image. 
}
\label{fig:first-stage-training}
\end{figure*}

In our experiments, we observe that the mixing network learns to ignore the $\text{W}^{rand}$.
To achieve diversity, we propose a training pipeline with GAN-generated data as shown in Fig. \ref{fig:first-stage-training}.
We first generate an image with $z_g$, and feed it to the mapping network to obtain $\text{W}^{rand}_g $. We synthesize the output image and erase a part of the image with a randomly generated mask,
$\text{I}_g^e = \text{M} \odot \text{I}_g$.
Our encoder encodes the erased image to $\text{W}^{enc}_g$.
With these encodings, we generate two images, one that combines $\text{W}^{enc}_g$ with mapping of $z_g$ and another one with mapping of a randomly sampled different $z_r$ as given in Fig. \ref{fig:first-stage-training} (a) and (b), respectively.
The generator outputs images $\text{I}_g^o$ and $\text{I}_r^o$ from these two different paths.
The generation of $\text{I}_g^o$ has access to the full information that generated the image and therefore, if one can combine the encoded embedding with the mapping, one correctly can generate the input image. 
On the other hand, the second path, $\text{I}_r^o$ image has only access to the information of unerased pixels and is not expected to reconstruct the input image faithfully for the erased parts. 
To achieve this objective, we set the following losses to train the framework.

\textbf{Reconstruction Losses.} For reconstructing the pixels correctly, we use  $L_2$ and perceptual losses.
We use perceptual losses from VGG ($\Phi$) at different feature layers ($j$) between images.
We output two images, $\text{I}_g^o$ and $\text{I}_r^o$ from the input image ($\text{I}_g$) that is erased, $\text{I}_g^e$.
$\text{I}_g^o$ has access to the correct mapping as well as unerased image pixels, therefore this training pipeline is expected to generate the original input image as follows:

\begin{equation}
    \begin{split}
        \mathcal{L}_{rg} = || {\text{I}_g^o} -  {\text{I}_g} ||_2 + |\Phi_{j}({\text{I}_g^o}) - \Phi_{j}({\text{I}_g} ) ||_2
    \end{split}
    \label{eqn:rec_g}
\end{equation}

On the other hand,  $\text{I}_r^o$ has only access to the unerased pixels of the original image and is expected to fill the missing information with a randomly sampled latent code. 
Therefore, we only expect pixel-wise matching on the unerased images. 
We find the reconstruction losses between the masked output image and erased input image as follows:

\begin{equation}
    \begin{split}
        \mathcal{L}_{rr} = || {(\text{M} \odot \text{I}_r^o)} - { \text{I}_g^e} ||_2 
        + |\Phi_{j}({\text{M} \odot \text{I}_r^o}) - \Phi_{j}({ \text{I}_g^e} ) ||_2
    \end{split}
    \label{eqn:rec_r}
\end{equation}

If the network is trained with Eq. \ref{eqn:rec_g} alone, the mixing network can ignore the input image and encoded features coming from it since full information is provided by the mapping network. Similarly, if the network is trained with only Eq. \ref{eqn:rec_r}, the mixing network ignores the mapping network.
By using these reconstruction losses together, our framework learns to use the information encoded from both paths.
Additionally, this training regularizes the encoder and mixing network to output latent codes that are in GAN's natural space which provides editable inversions.


\textbf{Adversarial Losses.} We additionally output our final images as follows:

\begin{equation}
 \begin{split}
    \text{I}^f = \text{M} \odot \text{I} + \text{(1-M)} \odot \text{I}^o
\end{split}
\label{eq:final}
\end{equation}

This step guarantees that the valid pixels are not modified as given in Fig. \ref{fig:first-stage-arch}.
We expect these final images to look realistic and use adversarial guidance on the final images for both $\text{I}^f_g$ and  $\text{I}^f_r$ .
We load the pretrained discriminator from StyleGAN training, $D$, and train the discriminator together with the encoder and mixing network.

\begin{equation}
 \begin{split}
    L_{adv} = 2\log{ D( {\text{I}_g} )} + \log{(1-D({\text{I}^f_g}))} \\ + \log{(1-D({\text{I}^f_r}))}
    \end{split}
\end{equation}

\textbf{Full Objective.}
We use the overall objectives given below to optimize the parameters of the Encoder (E) and the mixing network (Mi). The hyperparameters are provided in Appendix.

\begin{equation}
    \begin{split}
        \underset{E,Mi}{\min} \underset{D}{\max} \lambda_{a}\mathcal{L}_{adv} +  \lambda_{r1} \mathcal{L}_{rg} + \lambda_{r2} \mathcal{L}_{rr}   
    \end{split}
    \label{eqn:full_loss}
\end{equation}

\subsection{Second Stage: High-Fidelity Reconstruction}
\label{sec:second}

Previous works for image inversion show that low-rate latent codes, $W+$, do not have enough capacity for high-fidelity image reconstruction \cite{wang2022high, alaluf2022hyperstyle}. 
There is too much information loss due to the bottleneck of projecting high-dimensional images to low-rate latent codes and this bottleneck limits the high-fidelity reconstructions.
The same is observed by inpainting with image inversion works \cite{wang2022dual, yu2022high} as well.
The high-fidelity reconstruction of visible pixels is even more crucial in the inpainting task because we combine the erased and generated pixels as given in Eq. \ref{eq:final}.
When the generator is not able to reconstruct the valid pixels with high fidelity, the color discrepancies become very obvious in the final images.
Previous methods \cite{wang2022dual, yu2022high, wang2022high, alaluf2022hyperstyle} suggest encoding information to higher-rate latent codes as well. 
This is achieved by skip connections from the encoder to the generator on higher-resolution feature maps which are sometimes referred to as $F^+$ space.

The balance of encoding information between $W^+$ and $F^+$ spaces is extensively studied for image inversion methods \cite{wang2022high, alaluf2022hyperstyle, pehlivan2022styleres}. That is because by encoding most of the information to $F^+$ space, one can guarantee a high-fidelity reconstruction. However, since the image is not projected to  $W+$ properly, it will not be editable which is the main goal of many inversion methods.
That is the reason we adopt the two-stage training pipeline similar to state-of-the-art image inversion methods \cite{wang2022high, alaluf2022hyperstyle, pehlivan2022styleres}. 
In the first stage, our method tries to reconstruct and inpaint images only by the encoded low-rate codes, $W^+$.
Later, we reduce the information bottleneck by letting higher rate codes skipping from the input image to the generator.

\begin{figure*}
\centering
\scalebox{0.71}{
\addtolength{\tabcolsep}{-5pt}   
\begin{tabular}{ccccccccc}

\\
\interpfigt{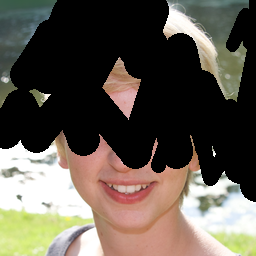} &
\interpfigt{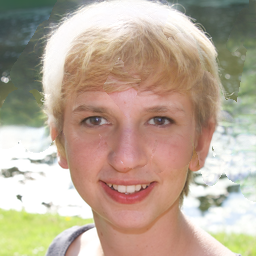} &
\interpfigt{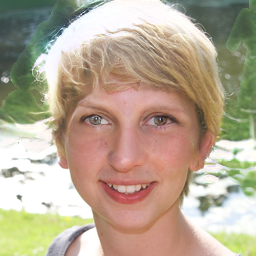} &
\interpfigt{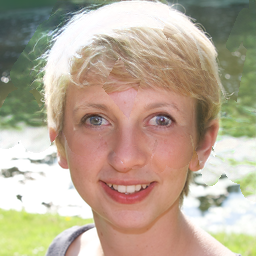} &
\interpfigt{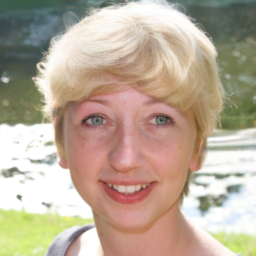} &
\interpfigt{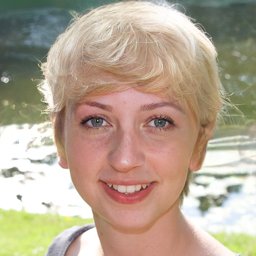} &
\interpfigt{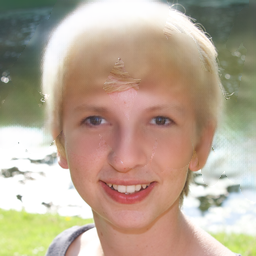} &
\interpfigt{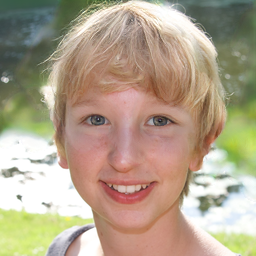} &
\interpfigt{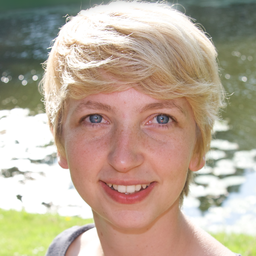}
\\
\interpfigt{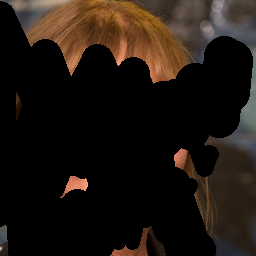} &
\interpfigt{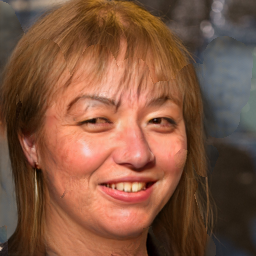} &
\interpfigt{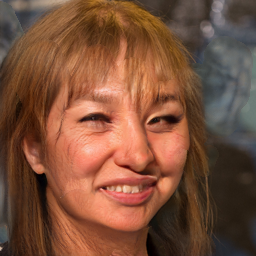} &
\interpfigt{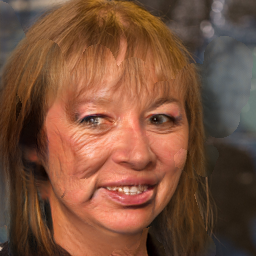} &
\interpfigt{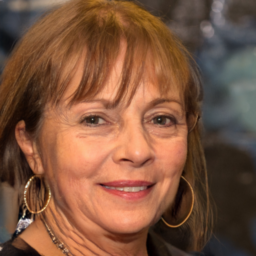} &
\interpfigt{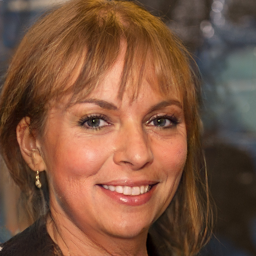} &
\interpfigt{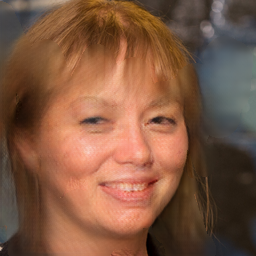} &
\interpfigt{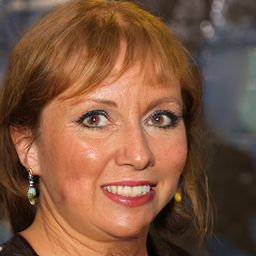} &
\interpfigt{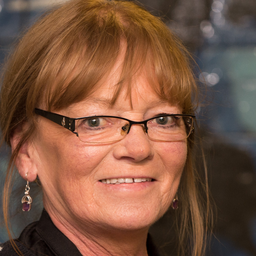}
\\
\interpfigt{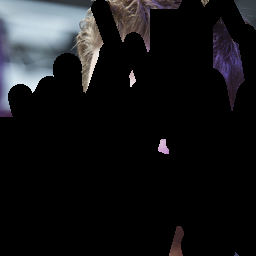} &
\interpfigt{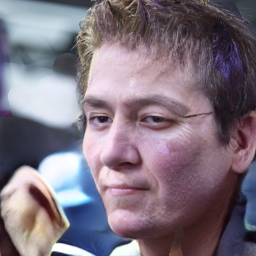} &
\interpfigt{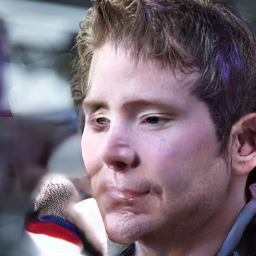} &
\interpfigt{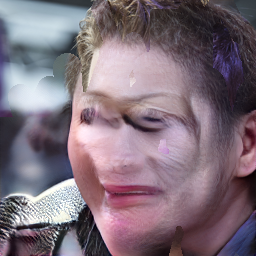} &
\interpfigt{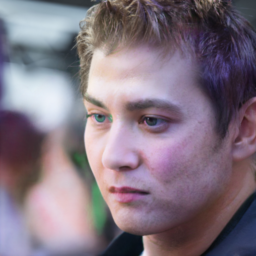} &
\interpfigt{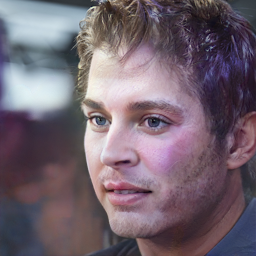} &
\interpfigt{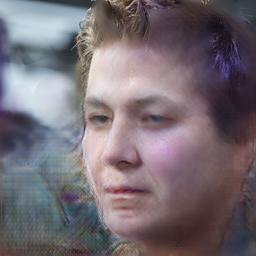} &
\interpfigt{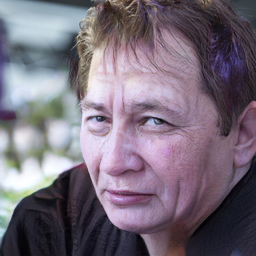} &
\interpfigt{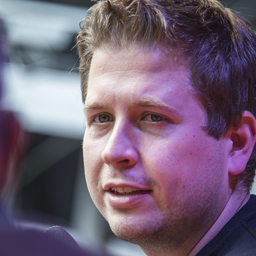}
\\
\interpfigt{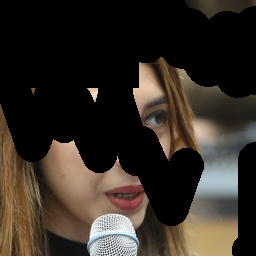} &
\interpfigt{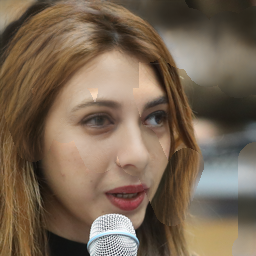} &
\interpfigt{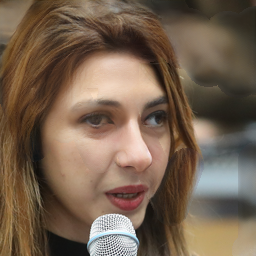} &
\interpfigt{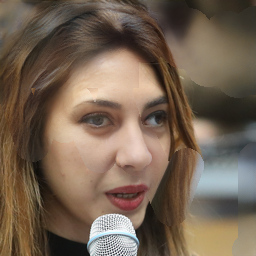} &
\interpfigt{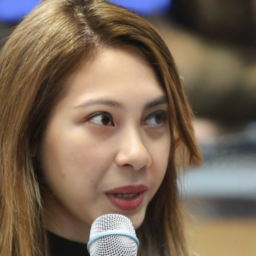} &
\interpfigt{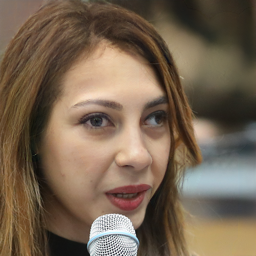} &
\interpfigt{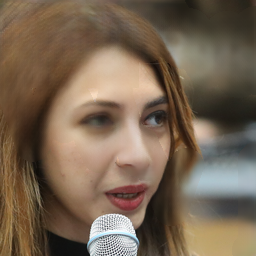} &
\interpfigt{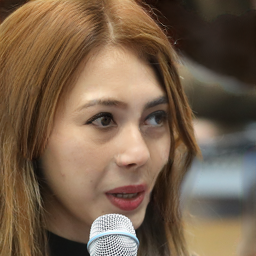} &
\interpfigt{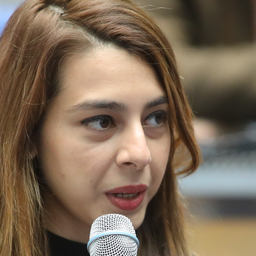}
\\
\interpfigt{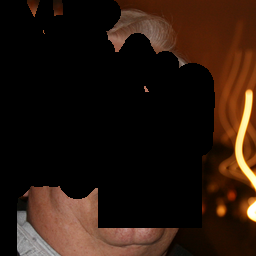} &
\interpfigt{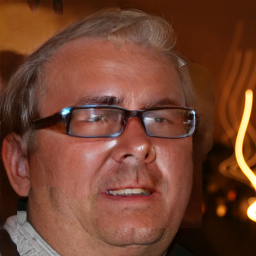} &
\interpfigt{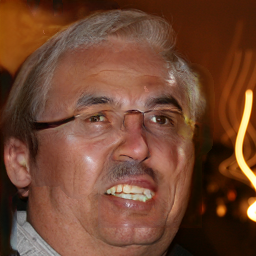} &
\interpfigt{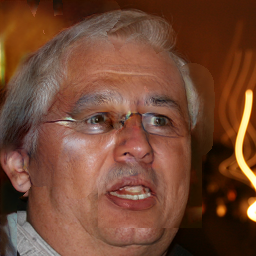} &
\interpfigt{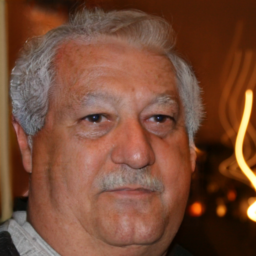} &
\interpfigt{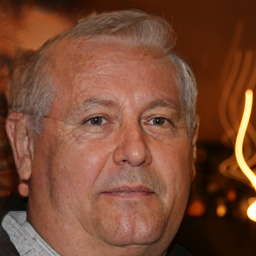} &
\interpfigt{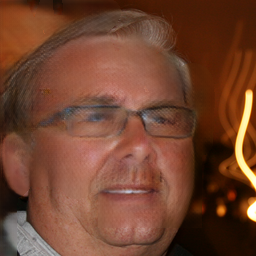} &
\interpfigt{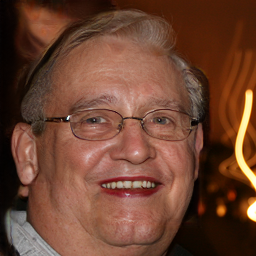} &
\interpfigt{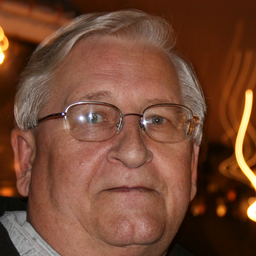}
\\
\interpfigt{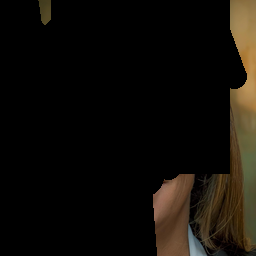} &
\interpfigt{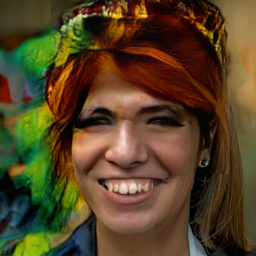} &
\interpfigt{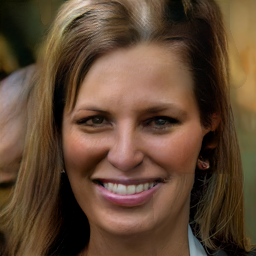} &
\interpfigt{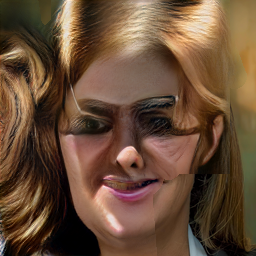} &
\interpfigt{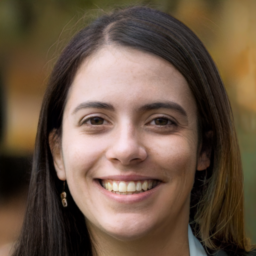} &
\interpfigt{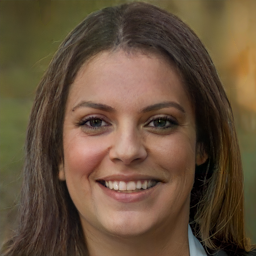} &
\interpfigt{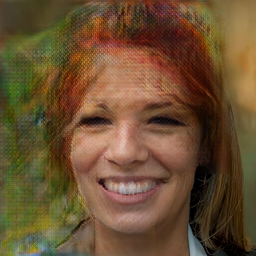} &
\interpfigt{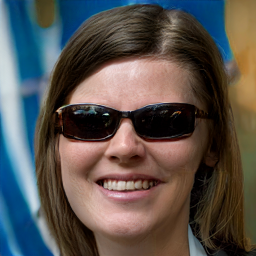} &
\interpfigt{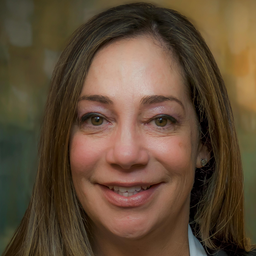}
\\
\interpfigt{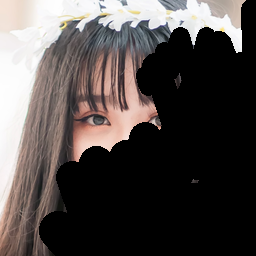} &
\interpfigt{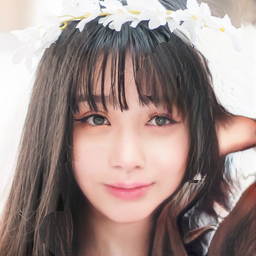} &
\interpfigt{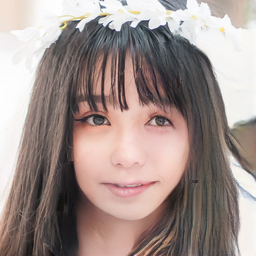} &
\interpfigt{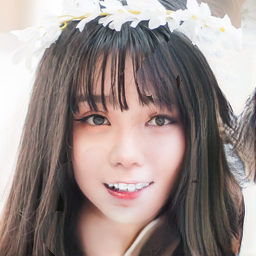} &
\interpfigt{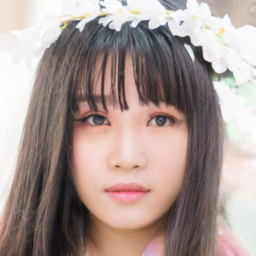} &
\interpfigt{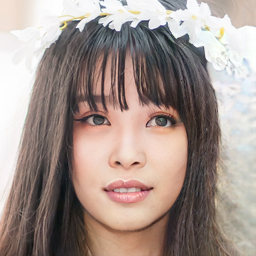} &
\interpfigt{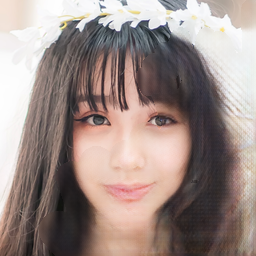} &
\interpfigt{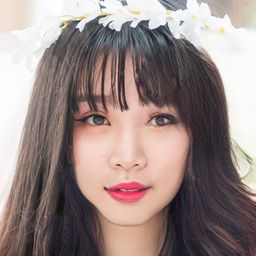} &
\interpfigt{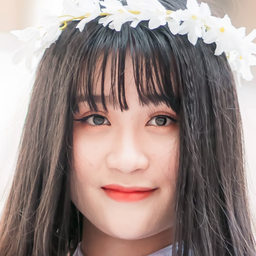}
\\
\interpfigt{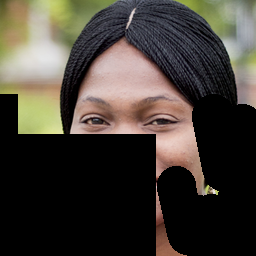} &
\interpfigt{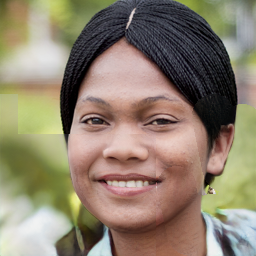} &
\interpfigt{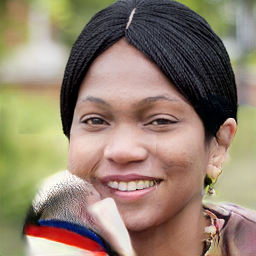} &
\interpfigt{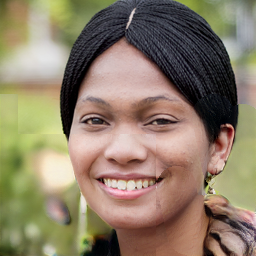} &
\interpfigt{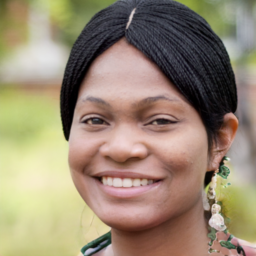} &
\interpfigt{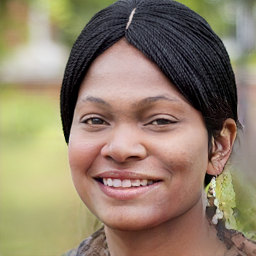} &
\interpfigt{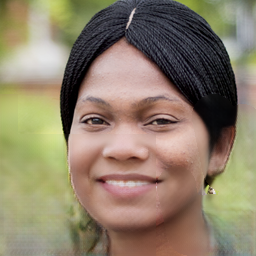} &
\interpfigt{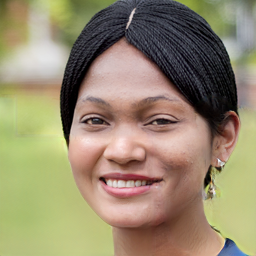} &
\interpfigt{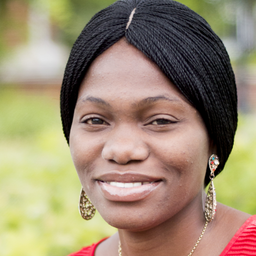}
\\
\interpfigt{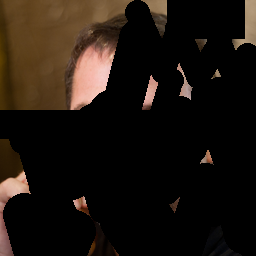} &
\interpfigt{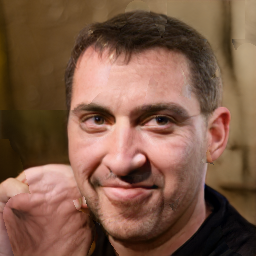} &
\interpfigt{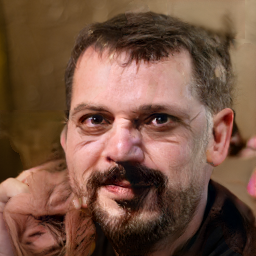} &
\interpfigt{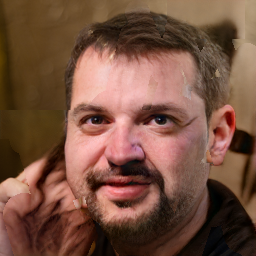} &
\interpfigt{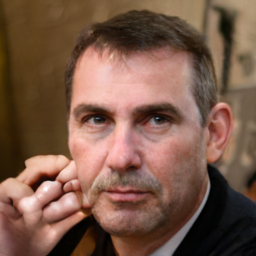} &
\interpfigt{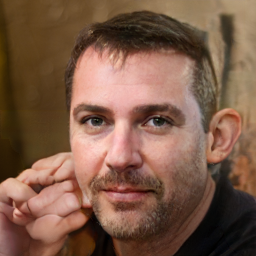} &
\interpfigt{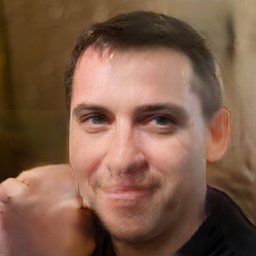} &
\interpfigt{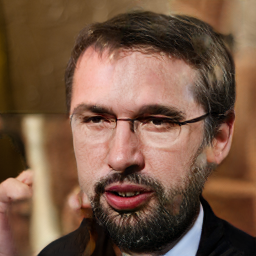} &
\interpfigt{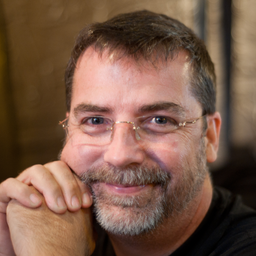}
\\
 Input &  pSp & HFGI & HyperStyle & CoModGAN  & InvertFill & DualPath & Ours & GT \\
\end{tabular}
}
\caption{Qualitative results of our and competing methods. GT refers to original images. }
\label{fig:results_all}
\end{figure*}

The second-stage training takes the final output of the first-stage model that has gone through the masking as given in Eq. \ref{eq:final}. We train an encoder with skip connections to the StyleGAN generator.
The encoder also takes the binary mask as input and can detect inconsistencies in the pixel level. 
The same encoded $W^+$ is used for both generations. 
However, in the first generation, there are no skip connections ($F^+$), and in the second stage, there are. 
We also limit the number of skip connection layers and only feed to the feature maps with resolutions of $32\times32$, $64\times64$, and $128\times128$.
We fine-tune the models with the same objective from the first stage, given in Eq. \ref{eqn:full_loss}.
We provide the architectural and training details in Appendix.

\section{Experiments}

\textbf{Baselines.} We first compare our method with state-of-the-art image inversion methods pSp \cite{richardson2021encoding}, HFGI \cite{wang2022high}, and HyperStyle \cite{alaluf2022hyperstyle}. 
We use the author's released code and train those models for inpainting tasks with an additional channel in the input for masks.
Note that this is a different comparison than the display in Fig. \ref{fig:teaser} since we train these models for inpainting.
pSp model outputs $W^+$ predictions for the generation. HFGI and HyperStyle methods use two-stage training. 
First, they rely on an encoder to output $W^+$ predictions, then a second encoder takes the input image and StyleGAN generated image with $W^+$ predictions. 
The goal is to encode the missed information to higher-rate latent codes.
We feed erased images and masks to these inversion models and train them to reconstruct the visible pixels faithfully and output realistic images overall.
Next, we experiment with state-of-the-art image inpainting models for our comparisons. 
We run inferences with CoModGAN's pretrained models \cite{zhao2021large}.
CoModGAN proposes to train a StyleGAN-like model from scratch but with co-modulation and skip connections for the inpainting task.
InvertFill \cite{yu2022high} and DualPath \cite{wang2022dual} models are built on pretrained StyleGAN models.
InvertFill's authors provide us with their inference results with the masks we provide. 
We implement DualPath following the set-up provided in their paper.
DualPath trains a generator that also takes StyleGAN features.

\textbf{Evaluation.}
We report Frechet Inception Distance (FID) metric \cite{heusel2017gans}, which looks at realism by comparing the target image distribution and reconstructed images.
We also look at the diversity scores with Learned Perceptual Image Patch Similarity (LPIPS) \cite{zhang2018unreasonable}.
That is for the models that provide diverse images, we generate two images per input image-mask pair and find the distance between these two images.
We additionally use U-IDS and P-IDS metrics that are proposed by CoModGAN \cite{zhao2021large}.
They measure unpaired and paired inception discriminative scores, respectively by measuring the linear separability in a pre-trained feature space.
Our goal is to output images that are not separable from real images. SVM is fitted to the data and we expect a higher error rate. 

\begin{table}[]
\centering
\caption{Quantitative results of our and competing methods on FFHQ validation dataset for mask ranges between 0-1. }
\resizebox{1.0\linewidth}{!}{
\begin{tabular}{|l|c|c|c|c|}
\hline
Models                & {FID}  \ \ $\Downarrow$   & LPIPS  \ \ $\Uparrow$  &  U-IDS $\Uparrow$ & P-IDS $\Uparrow$ \\ \hline
pSp \cite{richardson2021encoding}  & 8.23 & - & 14.05 & 6.60 \\
HFGI \cite{wang2022high} & 7.66 & - & 15.63 & 7.76 \\
HyperStyle \cite{alaluf2022hyperstyle} & 7.46 & - & 16.66 & 8.55\\
 \hline
 CoModGAN \cite{zhao2021large}  & 7.35 & 0.0790 &  9.78 & 3.79 \\ 
 InvertFill \cite{yu2022high} & 7.45 & - & 15.83 & 7.69  \\
 DualPath \cite{wang2022dual} & 17.60 & - & 4.43 & 0.77 \\
 \hline
 Ours & \textbf{5.92} & \textbf{0.2026} & \textbf{18.43} & \textbf{10.75}  \\ 
\hline
\end{tabular}}
\label{table:results_ffhq}
\end{table}

\textbf{Datasets.} We use FFHQ human face \cite{karras2019style}, AFHQ cat, and AFHQ dog image datasets \cite{choi2020stargan}. 
We use the StyleGAN2 pre-trained models \cite{karras2020analyzing} and use their train and validation splits.
We build our extensive comparison with other works on FFHQ human face dataset following previous works.

\begin{table}[]
\centering
\caption{Quantitative results of our and competing methods on AFHQ validation datasets for mask ranges between 0-1. }
\resizebox{1.0\linewidth}{!}{
\begin{tabular}{|l|c|c|c|c|}
\hline
& \multicolumn{2}{c|}{Dog} & \multicolumn{2}{c|}{Cat} \\
\hline
Models & {FID}  \ \ $\Downarrow$   & LPIPS  \ \ $\Uparrow$  & {FID}  \ \ $\Downarrow$   & LPIPS  \ \ $\Uparrow$  \\ \hline
pSp \cite{richardson2021encoding} & 26.458 & - & 14.262 & - \\
HFGI \cite{wang2022high} & 20.576 & - & 17.007 & - \\
HyperStyle \cite{alaluf2022hyperstyle} & 23.919 & - & 13.425 & - \\
\hline
Ours & \textbf{18.890} & \textbf{0.062} & \textbf{12.275} & \textbf{0.087} \\ 
\hline
\end{tabular}}
\label{table:results_afhq}
\end{table}


\begin{figure}[t]
    \centering
\includegraphics[width=1.0\linewidth]{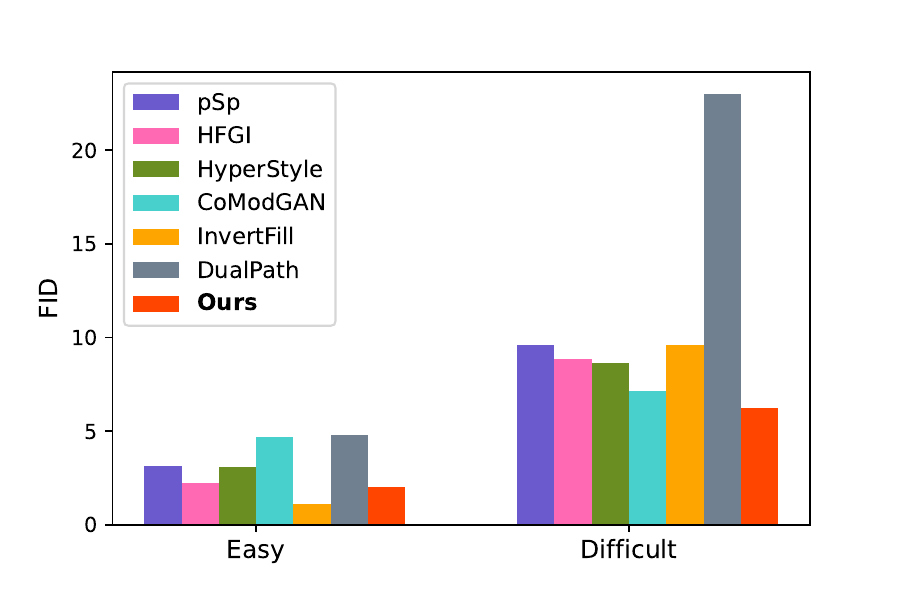}
    \caption{FID plots of ours and competing methods for Easy (0.0-0.4) and Difficult (0.4-1.0) mask ranges.}
    \label{fig:mask_plot}
\end{figure}

\begin{figure*}
\centering
\scalebox{0.6}{
\addtolength{\tabcolsep}{-5pt}   
\begin{tabular}{cccccccccc}
\\
\interpfigt{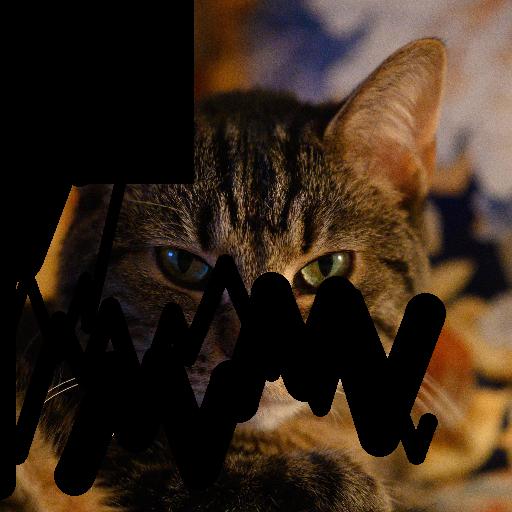} &
\interpfigt{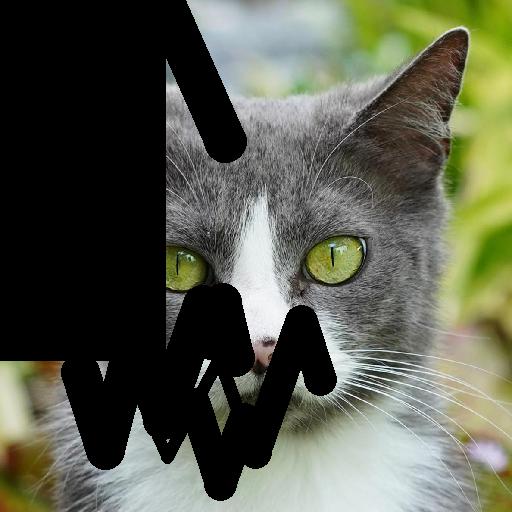} &
\interpfigt{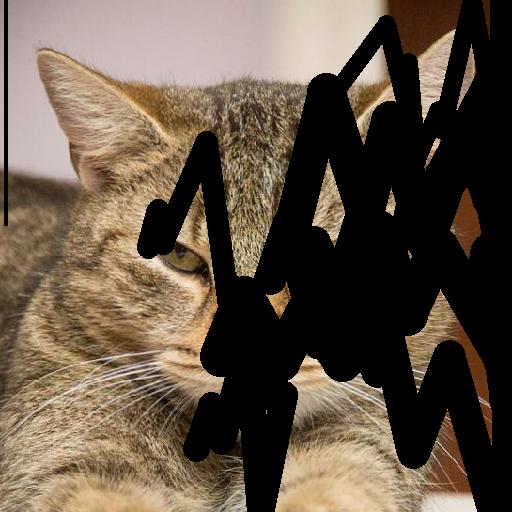} &
\interpfigt{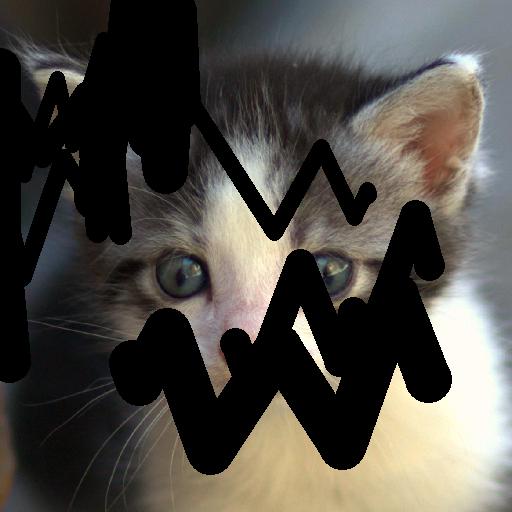} &
\interpfigt{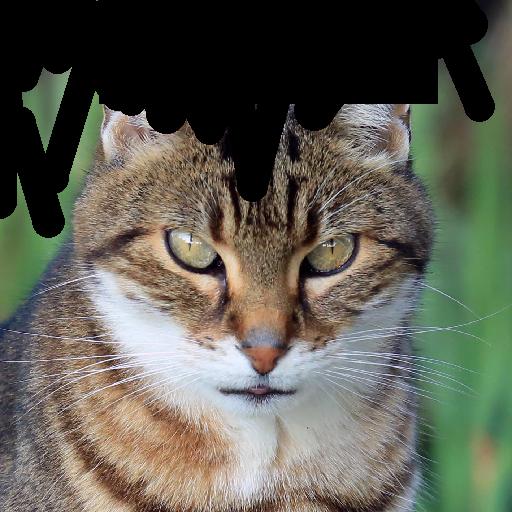} &
\interpfigt{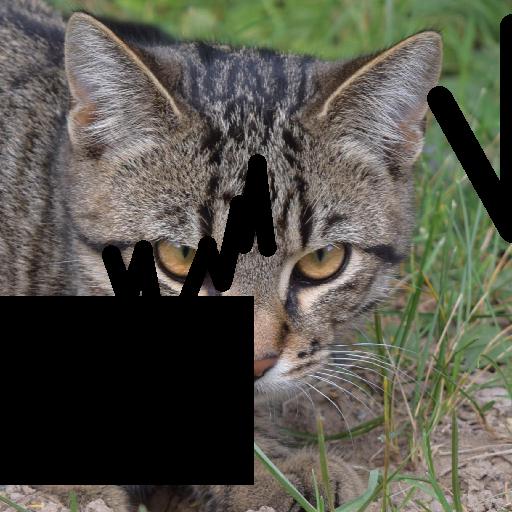} &
\interpfigt{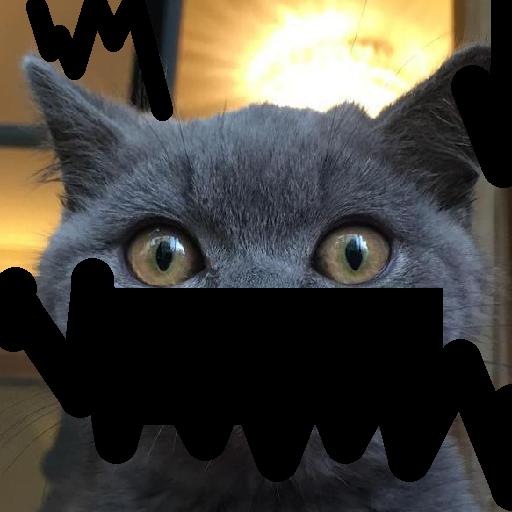} &
\interpfigt{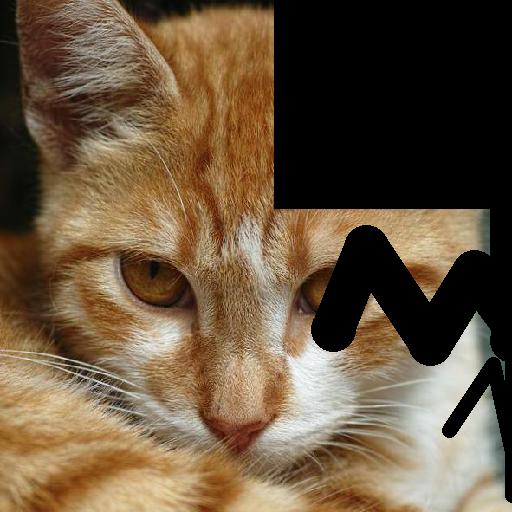} &
\interpfigt{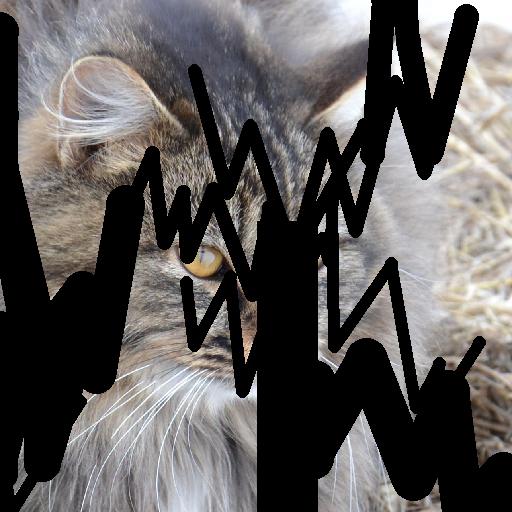} &
\interpfigt{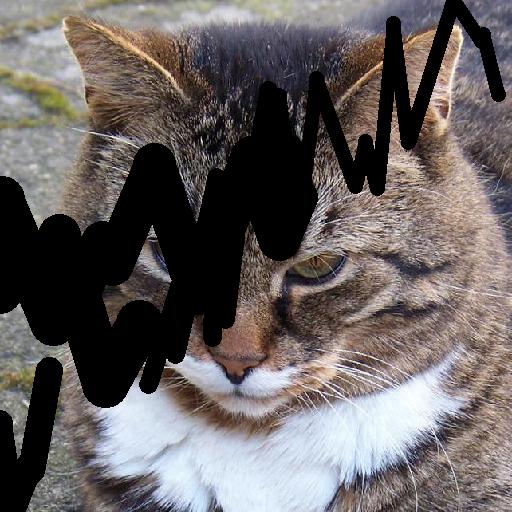}
\\

\interpfigt{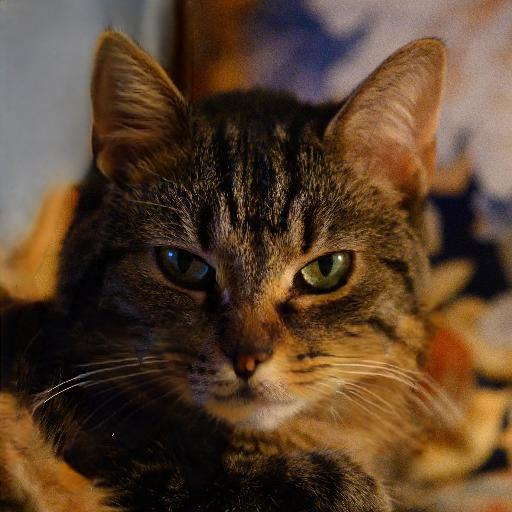} &
\interpfigt{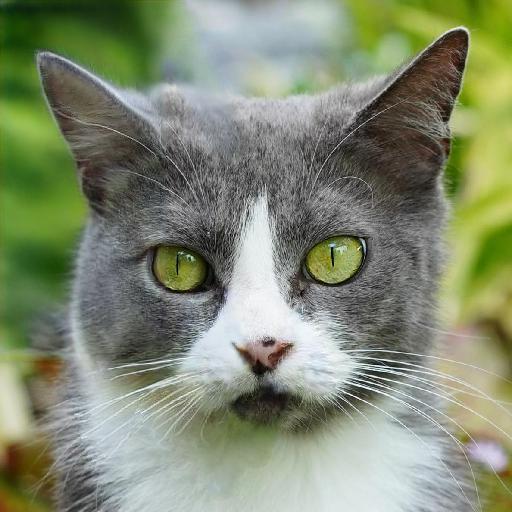} &
\interpfigt{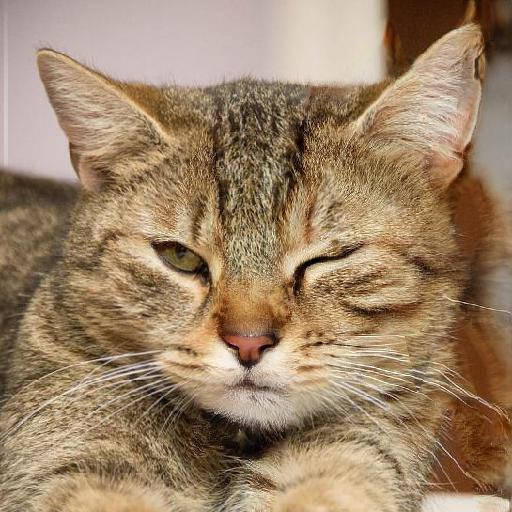} &
\interpfigt{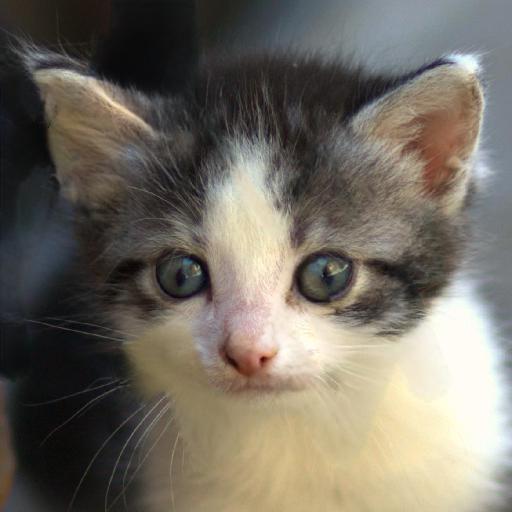} &
\interpfigt{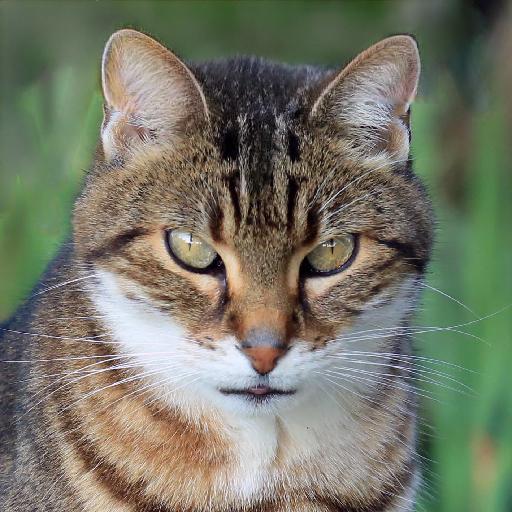} &
\interpfigt{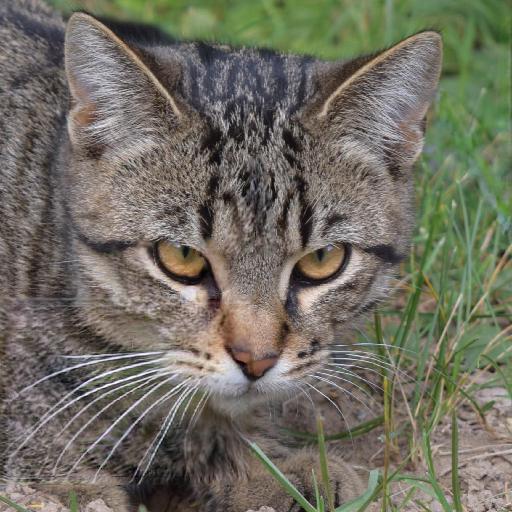} &
\interpfigt{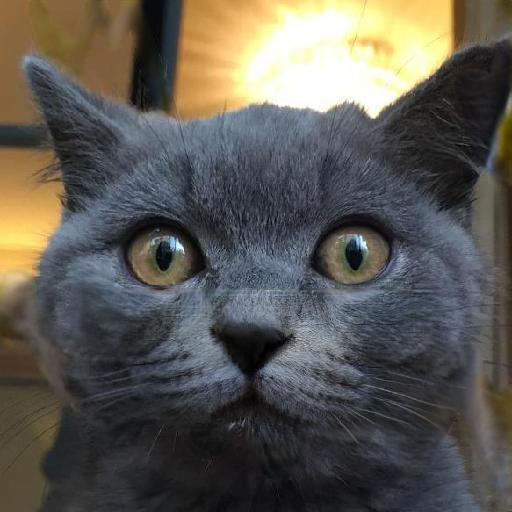} &
\interpfigt{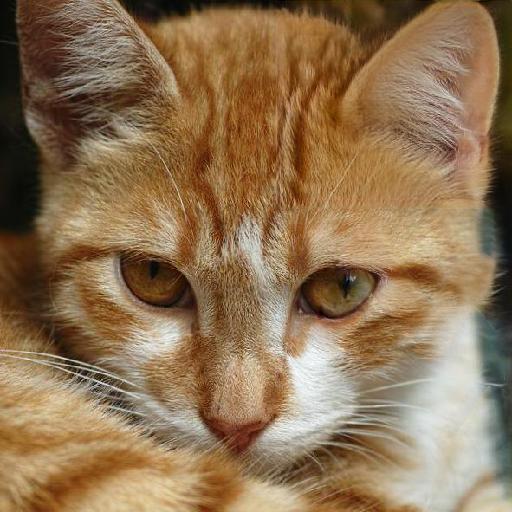} &
\interpfigt{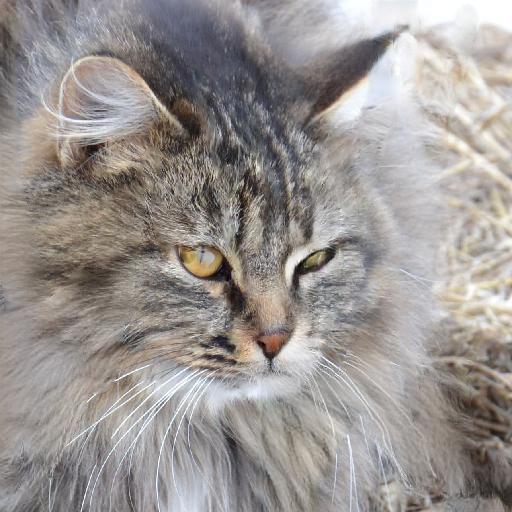} &
\interpfigt{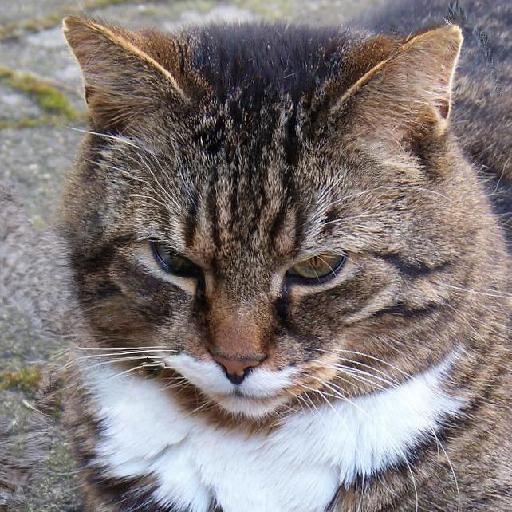}
\\
\interpfigt{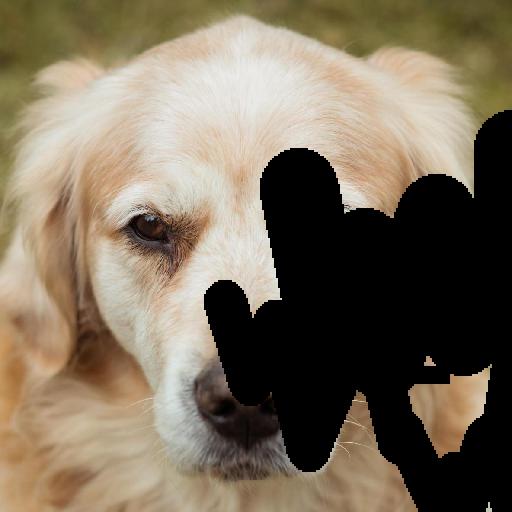} &
\interpfigt{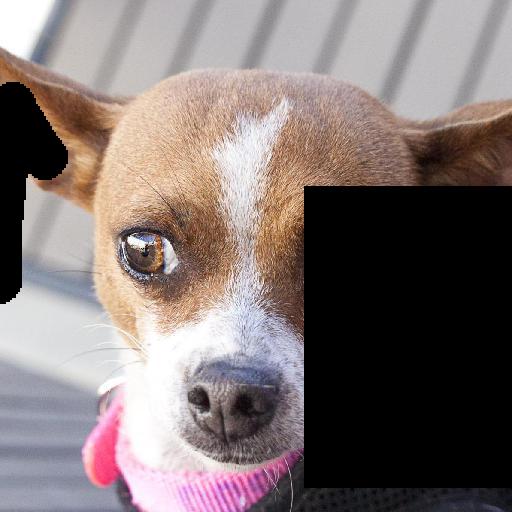} &
\interpfigt{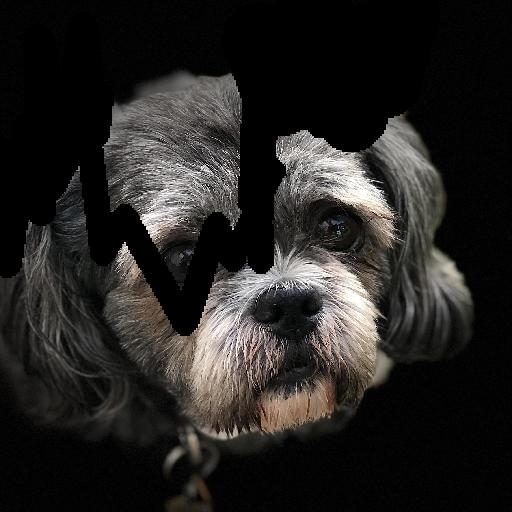} &
\interpfigt{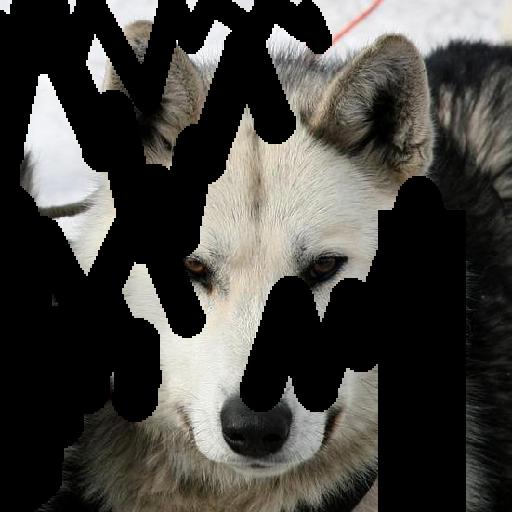} &
\interpfigt{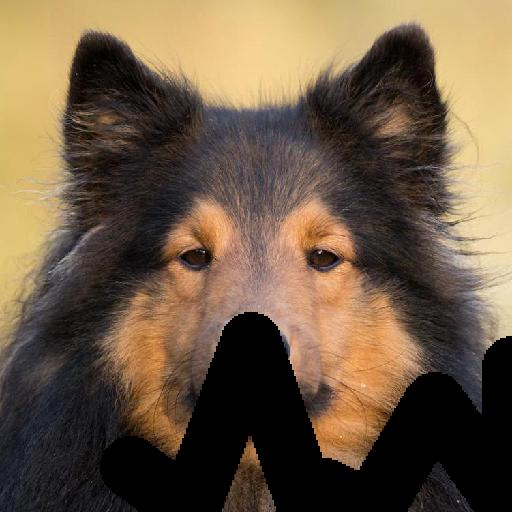} &
\interpfigt{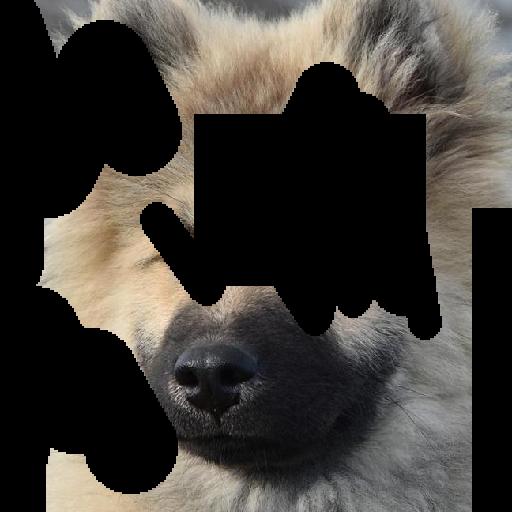} &
\interpfigt{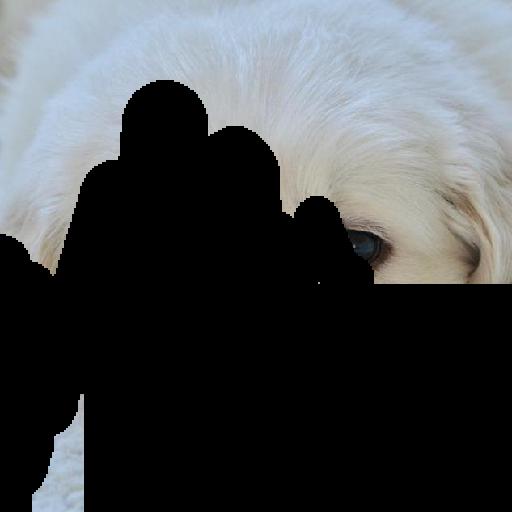} &
\interpfigt{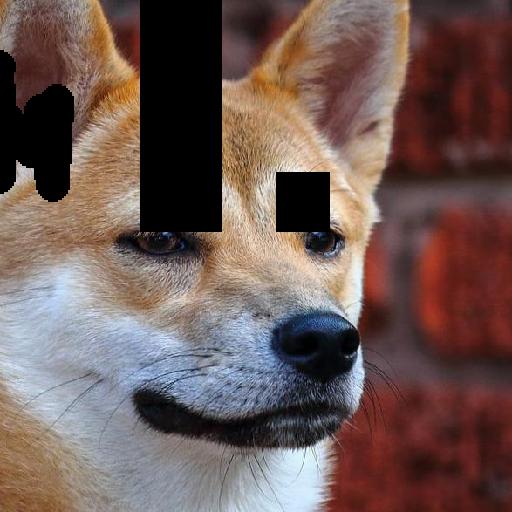} &
\interpfigt{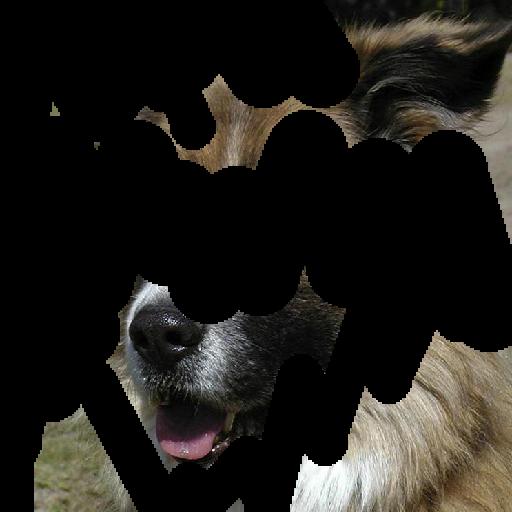} &
\interpfigt{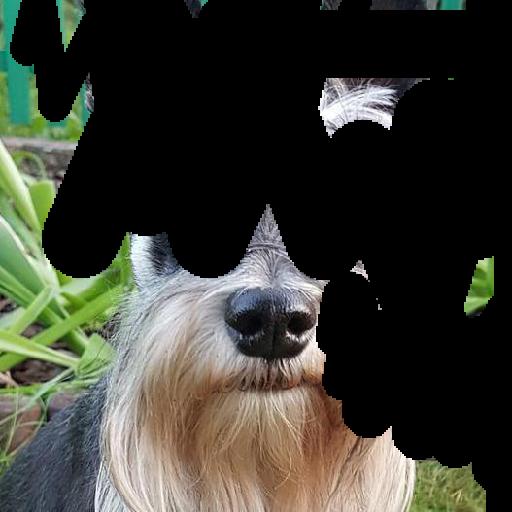}
\\
\interpfigt{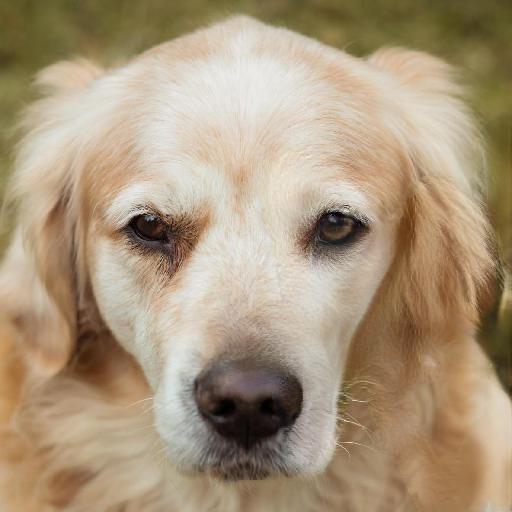} &
\interpfigt{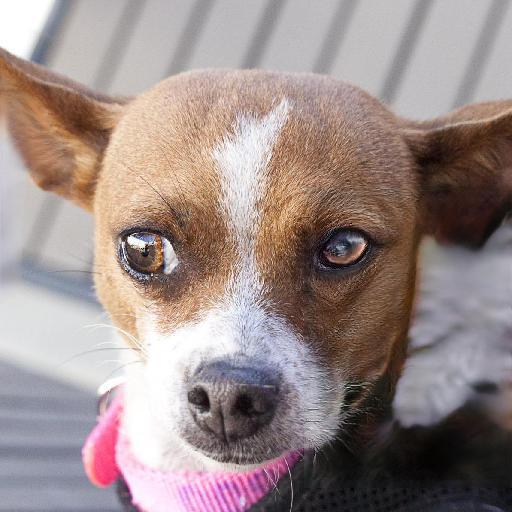} &
\interpfigt{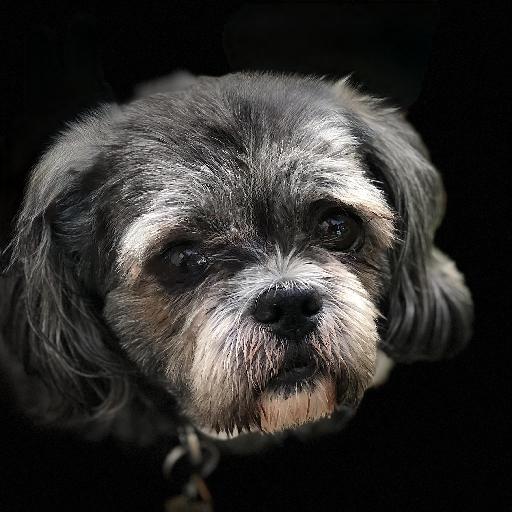} &
\interpfigt{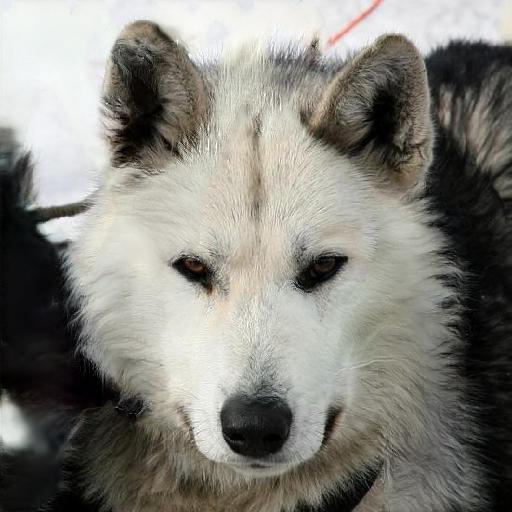} &
\interpfigt{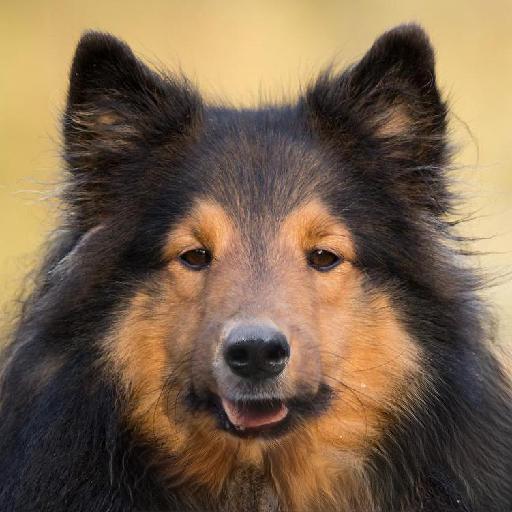} &
\interpfigt{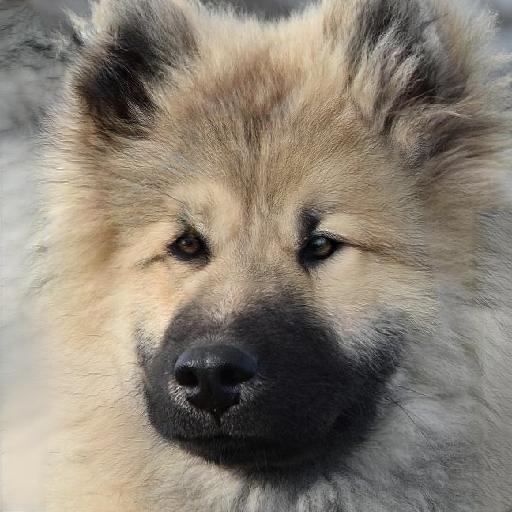} &
\interpfigt{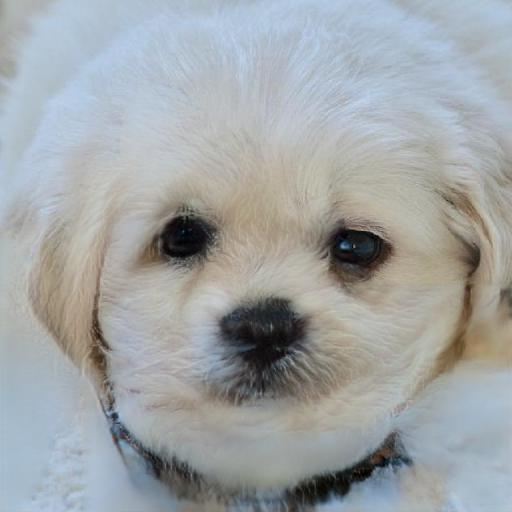} &
\interpfigt{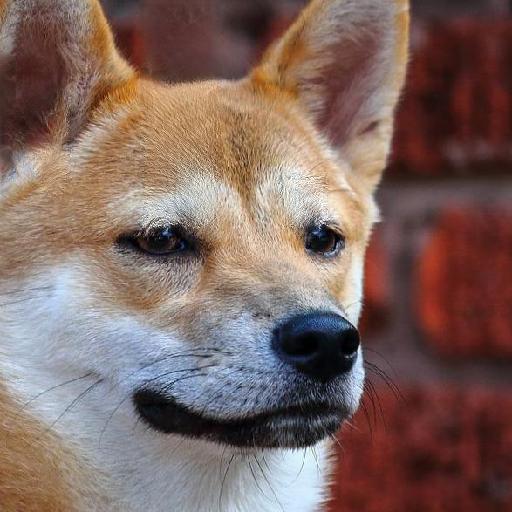} &
\interpfigt{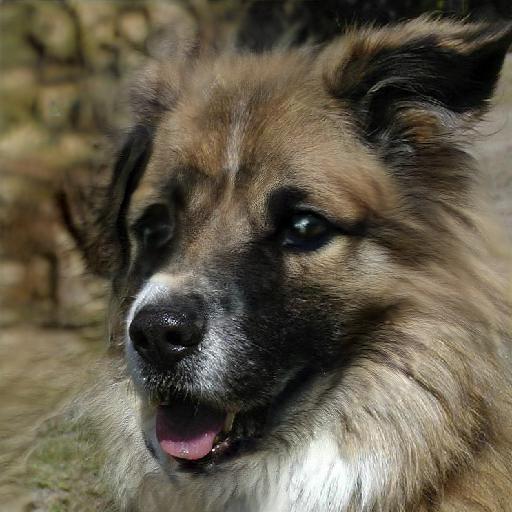} &
\interpfigt{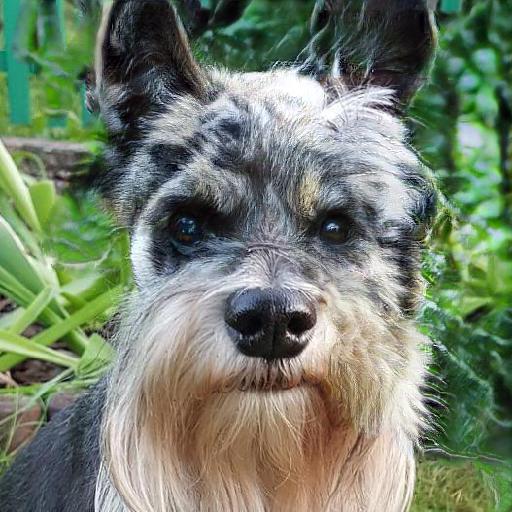}
\\
\end{tabular}
}
\caption{Qualitative results of AFHQ Cat and Dog dataset trainings.}
\label{fig:cat}
\end{figure*}

\begin{figure}
\centering
\scalebox{0.6}{
\addtolength{\tabcolsep}{-5pt}   
\begin{tabular}{ccccccc}
\\
\rotatebox{90}{~~~~~~~ID 1} & \interpfigt{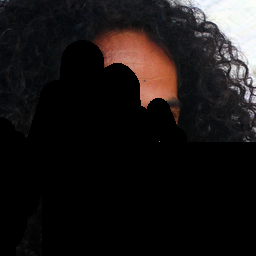} &
\interpfigt{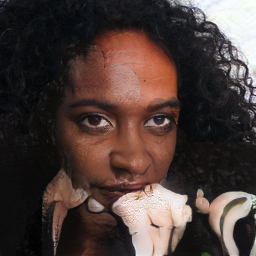} &
\interpfigt{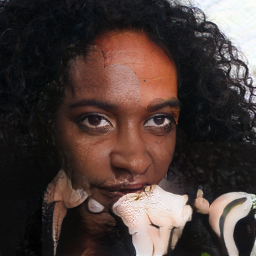} &
\interpfigt{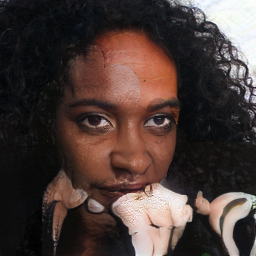} &
\interpfigt{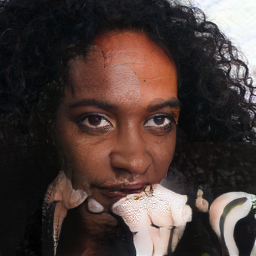} \\
\rotatebox{90}{~~~~~~~ ID 2} & \interpfigt{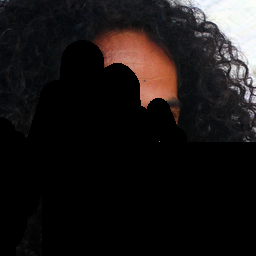} &
\interpfigt{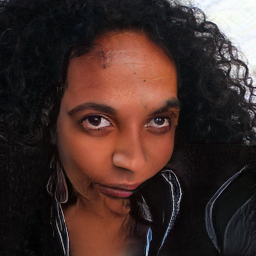} &
\interpfigt{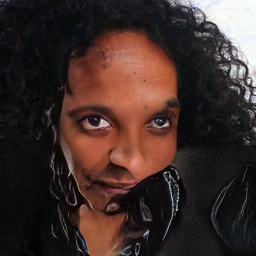} &
\interpfigt{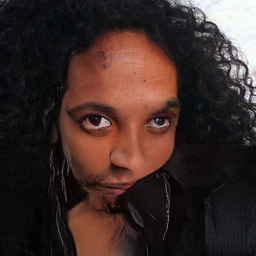} &
\interpfigt{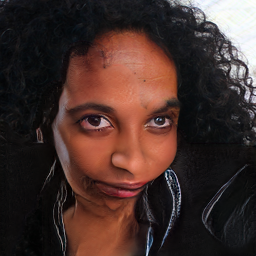} \\
\rotatebox{90}{~~~~~~~ ID 3} & \interpfigt{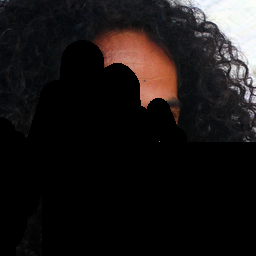} &
\interpfigt{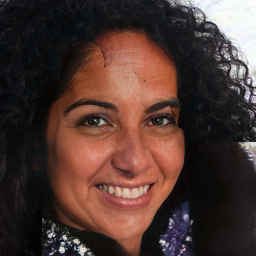} &
\interpfigt{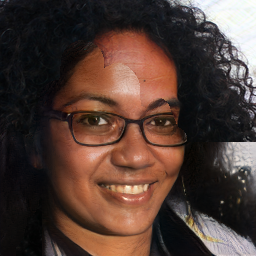} &
\interpfigt{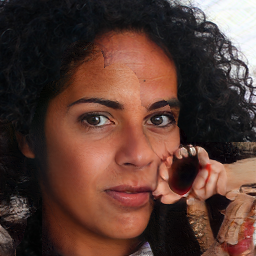} &
\interpfigt{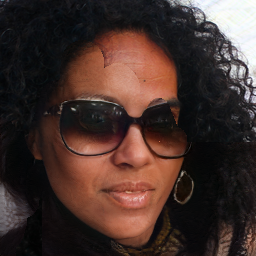} \\
\rotatebox{90}{~~~~~~~ ID 4} & \interpfigt{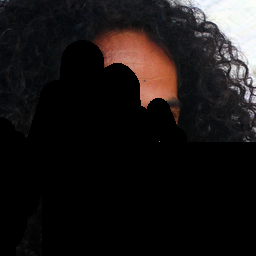} &
\interpfigt{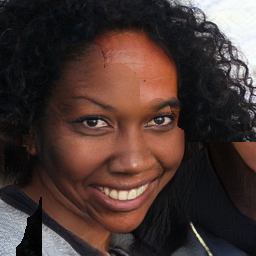} &
\interpfigt{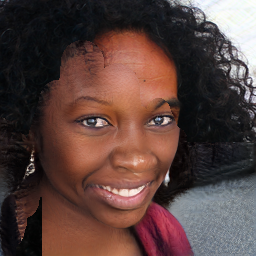} &
\interpfigt{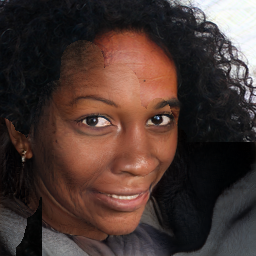} &
\interpfigt{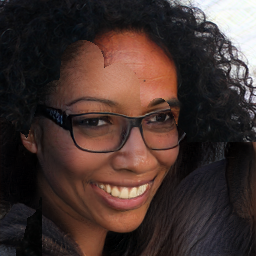} \\
\rotatebox{90}{~~~~~~~ ID 5} & \interpfigt{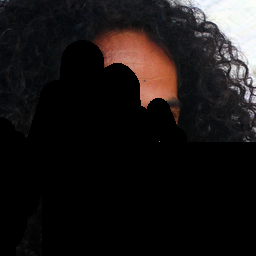} &
\interpfigt{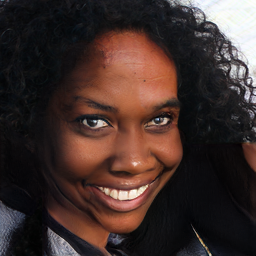} &
\interpfigt{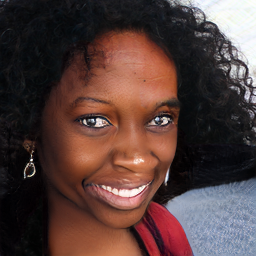} &
\interpfigt{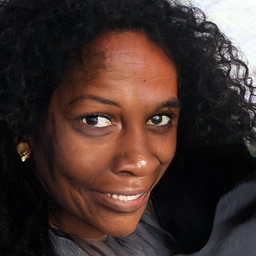} &
\interpfigt{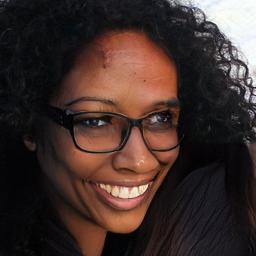} \\
\end{tabular}
}
\caption{Qualitative results of the ablation study. Please refer to Table \ref{table:results_ablation} for the training-setting of model IDs.}
\label{fig:ablation_div}
\end{figure}

\begin{table}[t]
\caption{Quantitative results of our ablation study on FFHQ validation dataset for the mask of range 0-1.
We assign model IDs to easily refer to the model in the text and Fig. \ref{fig:ablation_div}.}
\centering
\resizebox{\linewidth}{!}{
\begin{tabular}{|c|c|c|c|c|c|}
\hline
ID &   Full Recons & Gated Mi & Second Stage  &  FID   & LPIPS   \\ \hline
1 &  &  \checkmark & & 10.92 & 0.0073 \\
2&  & \checkmark &  \checkmark & 7.87 & 0.0855 \\
3&   \checkmark & & &  {20.91} & 0.2095 \\
4&    \checkmark &  \checkmark & & {16.65} & 0.1440 \\
 5&  \checkmark &  \checkmark &  \checkmark & 5.92 & 0.2026 \\
\hline
\end{tabular}}
\label{table:results_ablation}
\end{table}


\textbf{Quantitative Results.} 
We provide the quantitative results in Table \ref{table:results_ffhq}, \ref{table:results_afhq}.
We provide the LPIPS score for the models that provide diversity. 
The inversion methods specifically pSp, HFGI, and HyperStyle, that are trained for inpainting task achieves reasonable scores.
Next, we compare with image inpainting methods. 
CoModGAN trains a network from scratch for inpainting whereas InvertFill uses a pretrained StyleGAN similar to our work.
CoModGAN has a StyleGAN-like training set-up, and layers are co-modulated with sampling and encoded image features.
CoModGAN achieves diversity but is lower than our model when measured with the LPIPS score.
InvertFill, on the other hand, does not have a mechanism to provide any stochastic at all.
They achieve similar FID scores but InvertFill achieves significantly better U-IDS and P-IDS scores.
We also compare with DualPath which is also built on pretrained StyleGAN. 
DualPath trains a new generator that takes input from StyleGAN feature maps. 
We follow the setup provided in their paper and try our best on parameter tuning. 
However, we do not achieve a good FID score with DualPath. Specifically, the generated images look smooth and semantically consistent, but they do not look sharp. 
DualPath also does not provide diverse results.
We achieve significantly better results than previous methods and also achieve diversity.

We analyze how different methods score with different difficulties of masks in Fig. \ref{fig:mask_plot}.
We set two difficulty levels, an easy one where the erased mask ratio is between (0, 0.4) and a difficult scenario where the erased mask ratio is between (0.4-1.0).
When the erased area only contains a small portion, deterministic models also achieve good results since most of the information can be encoded from the unerased parts.
InvertFill achieves even a better score than ours and they all achieve better than CoModGAN.
However, when the difficulty level increases, deterministic models struggle. 
CoModGAN, the other model that can output diverse results achieves the second-best results. 
InvertFill's FID score significantly increase. 
Our model achieves good scores in both scenarios.

\begin{figure*}
\centering
\scalebox{0.7}{
\addtolength{\tabcolsep}{-5pt}   
\begin{tabular}{ccccccccccccc}
\\
\rotatebox{90}{~~~~~~~Input} & 
\interpfigt{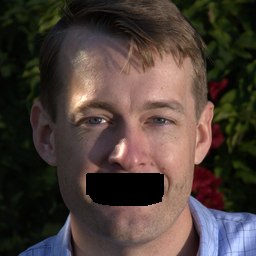} &
\interpfigt{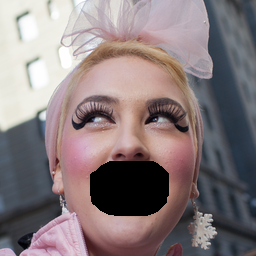} &
\interpfigt{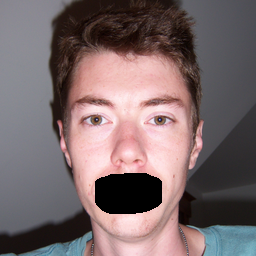} &
\rotatebox{90}{~~~~~~~Input} & 
\interpfigt{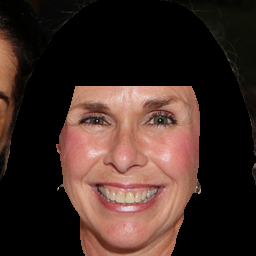} &
\interpfigt{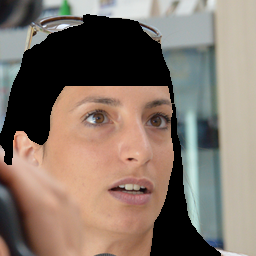} &
\interpfigt{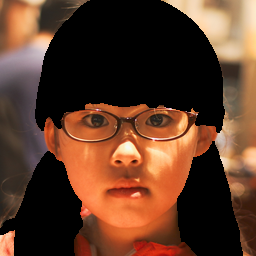} &
\rotatebox{90}{~~~~~~~Input} & 
\interpfigt{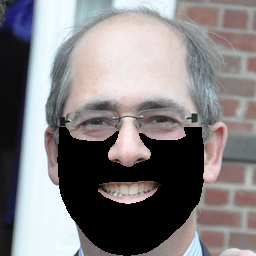} &
\interpfigt{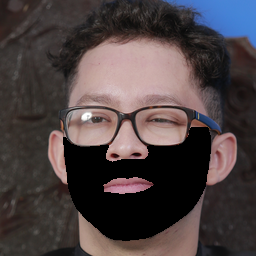} &
\interpfigt{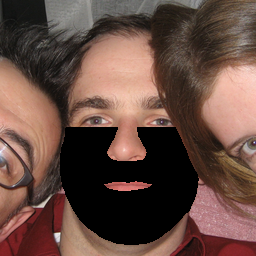} & 
\\
\rotatebox{90}{~~~~~~~Output} & 
\interpfigt{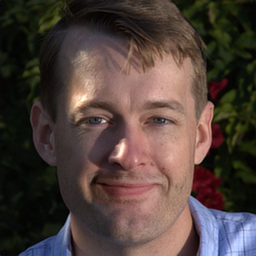} &
\interpfigt{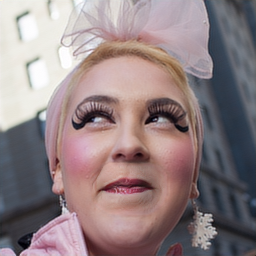} &
\interpfigt{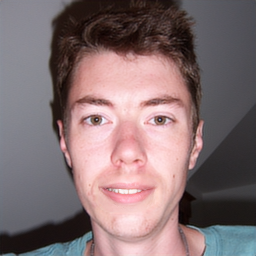} &
\rotatebox{90}{~~~~~~~Output}  & 
\interpfigt{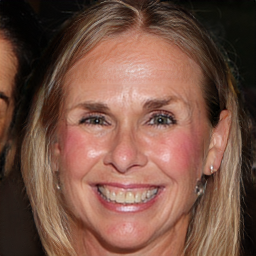} &
\interpfigt{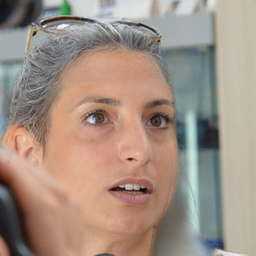} &
\interpfigt{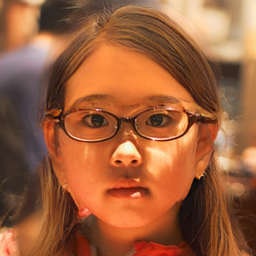} &
\rotatebox{90}{~~~~~~~Output}  & 
\interpfigt{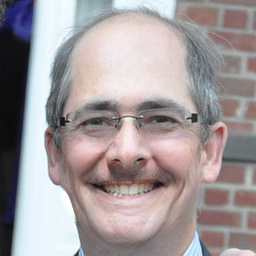} &
\interpfigt{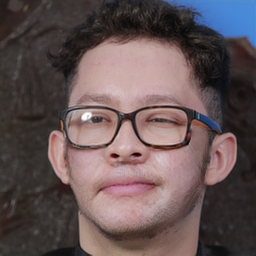} &
\interpfigt{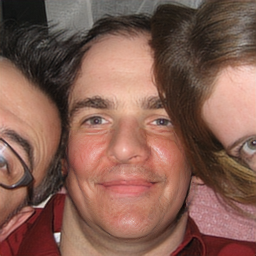} &  
\\
\rotatebox{90}{~~~~~~~Smile} & 
\interpfigt{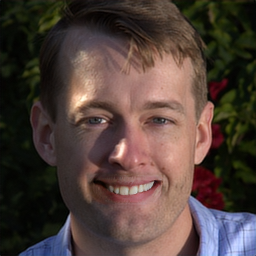} &
\interpfigt{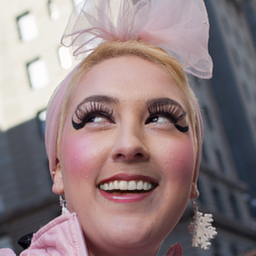} &
\interpfigt{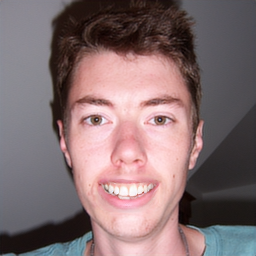} &
\rotatebox{90}{~~~~~~~Bangs}  & 
\interpfigt{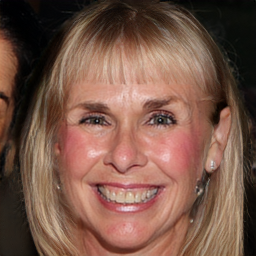} &
\interpfigt{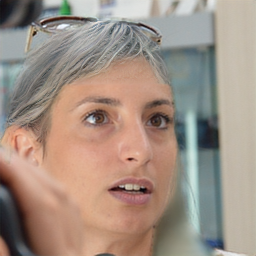} &
\interpfigt{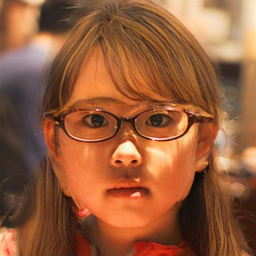} &
\rotatebox{90}{~~~~~~~~Beard}  & 
\interpfigt{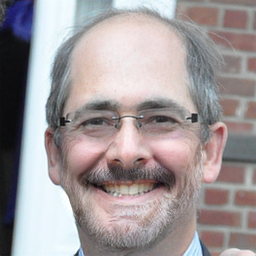} &
\interpfigt{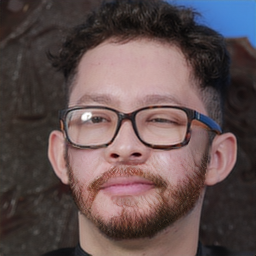} &
\interpfigt{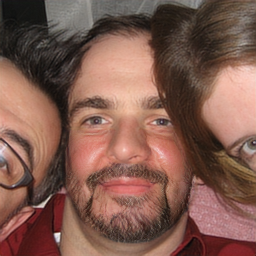} & 
\\
\rotatebox{90}{~~~~~~~Input} & 
\interpfigt{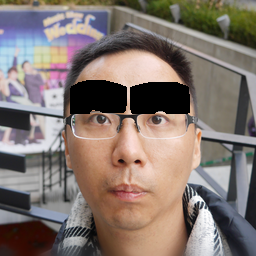} &
\interpfigt{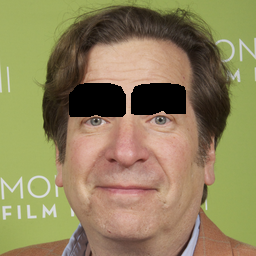} &
\interpfigt{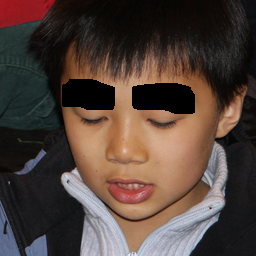} &
\rotatebox{90}{~~~~~~~Input} & 
\interpfigt{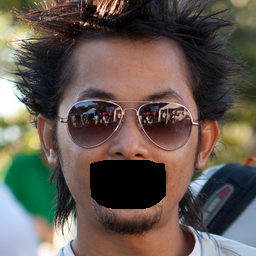} &
\interpfigt{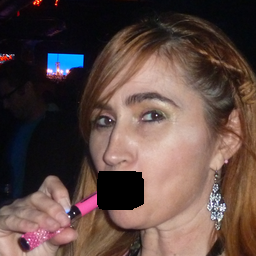} &
\interpfigt{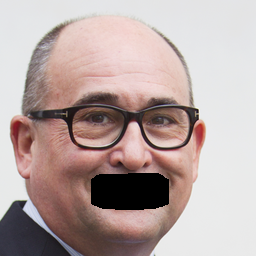} &
\rotatebox{90}{~~~~~~~Input} & 
\interpfigt{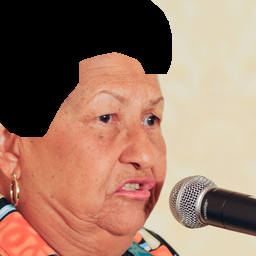} &
\interpfigt{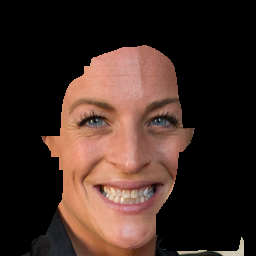} &
\interpfigt{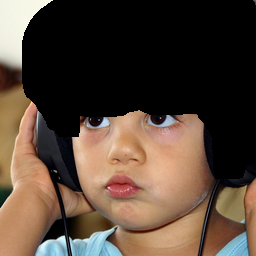} & 
\\
\rotatebox{90}{~~~~~~~Output} & 
\interpfigt{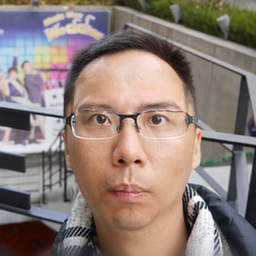} &
\interpfigt{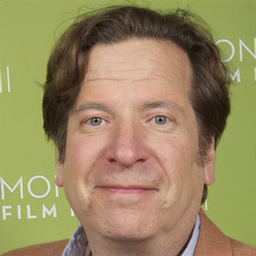} &
\interpfigt{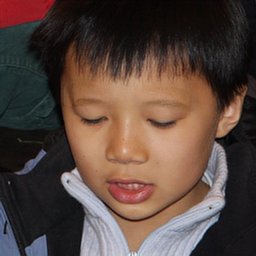} &
\rotatebox{90}{~~~~~~~Output}  & 
\interpfigt{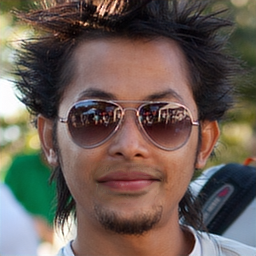} &
\interpfigt{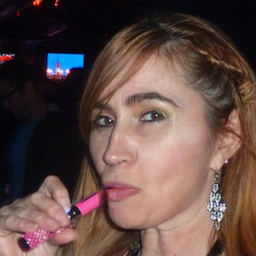} &
\interpfigt{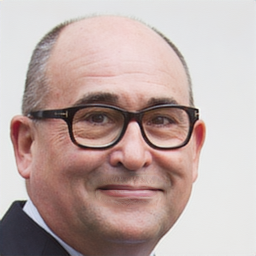} &
\rotatebox{90}{~~~~~~~Output}  & 
\interpfigt{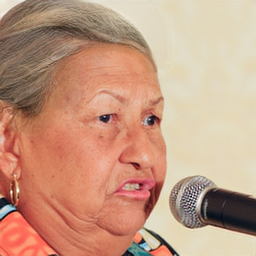} &
\interpfigt{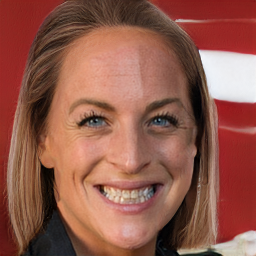} &
\interpfigt{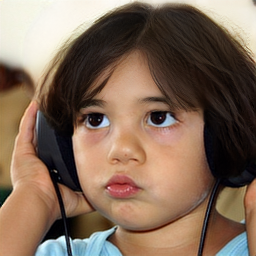} & 
\\
\rotatebox{90}{~~~~~~~Eyebrow} & 
\interpfigt{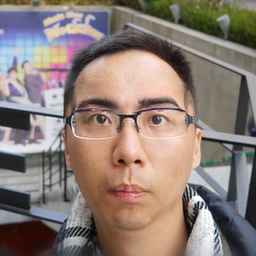} &
\interpfigt{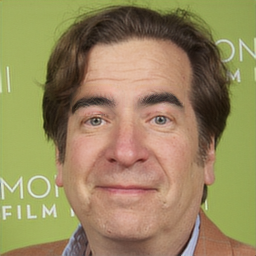} &
\interpfigt{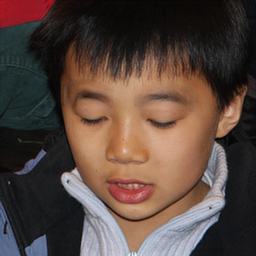} &
\rotatebox{90}{~~~~~~~Lipstick}  & 
\interpfigt{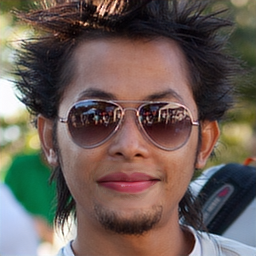} &
\interpfigt{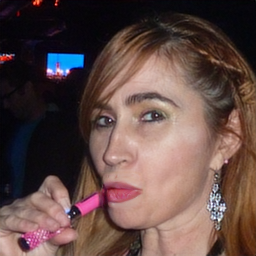} &
\interpfigt{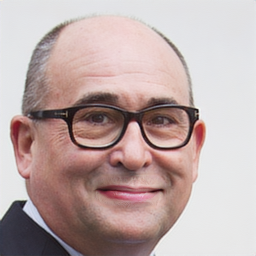} &
\rotatebox{90}{~~~~~~~Blonde}  & 
\interpfigt{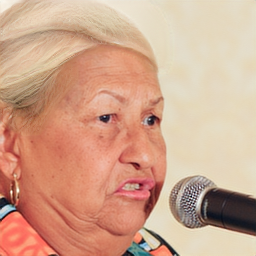} &
\interpfigt{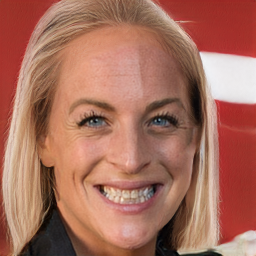} &
\interpfigt{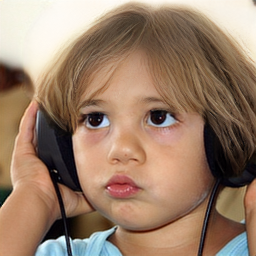} &  
\\

\end{tabular}
}
\caption{Editing results of our method. Our method achieves diverse inpainting and editing under one framework.}
\label{fig:editing_results}
\end{figure*}

\textbf{Qualitative Results.} Fig. \ref{fig:results_all} shows the results of our and competing methods on the FFHQ dataset. 
As we mention in the Quantitative Results section,  DualPath outputs blurry parts on the inpainted areas in our trainings.
The pSp model achieves good semantic consistency.
However, the boundaries between the inpainted and original pixels are quite apparent.
HFGI and HyperStyle output artifacts on many of the examples.
CoModGAN and InvertFill achieve plausible predictions.
An interesting observation we make is that all other methods output similar identities for given inputs. For example, the last two row predictions of CoModGAN and InvertFill are similar to each other. 
On the other hand, since our model samples diverse predictions (more diverse than CoModGAN), our results look quite different. Our results better capture the data distribution as is also measured by FID scores.
We also present the inpainting results of our method on AFHQ cat and dog images in Fig. \ref{fig:cat}.
Our method achieves successful inpaints on these image domains as well.

\textbf{Ablation Study.} We present the quantitative and qualitative results of our ablation study in Table \ref{table:results_ablation} and Fig. \ref{fig:ablation_div}, respectively.
During the development of our model, we track the diversity (LPIPS) and quality (FID) of our results.
LPIPS score alone does not reflect the quality of our results because random images can be generated by StyleGAN that are semantically inconsistent with valid pixels of the input image. High LPIPS scores can be obtained from those images.
In fact, during training, the LPIPS score starts from a high point and decreases as the framework learns to output consistent pixels with the unerased parts.
Therefore, it is important to achieve a reasonable FID and LPIPS together.


We assign IDs to each model in  Table \ref{table:results_ablation} to easily refer to them in the text and figures.
Model 1 is trained with real images to reconstruct the unerased parts together with adversarial loss.
The objective does not have the full-reconstruction image loss based on the sampled $W+$.
The model does not achieve any diversity even though the network architecture takes sampled $W+$ via the mixing network. 
The mixing network ignores the stochastic input as can be seen from Fig. \ref{fig:ablation_div}. 
In Model 2, we also add the second stage training which improves the FID results because it removes the color inconsistencies.
The diversity is negligible as can be seen from both Fig. \ref{fig:ablation_div} and Table \ref{table:results_ablation}.
Next, we compare methods that are trained with our proposed losses and training pipeline. 
Model 3 does not have the gated mixing network, instead, it employs a fully connected layer to combine encoded and sampled latent codes.
Model 4 has the gated mixing network as given in Eq. \ref{eq:gating}. 

In our experiments, we find that the proposed mixing network does a better job of combining features. 
Model 3 sometimes generates inconsistent results as can be seen in Fig. \ref{fig:ablation_div} and achieves worse FID scores.
For example, the input image may have half-erased eyeglasses and this network may not output eyeglasses for the visible parts. The gated mixing network achieves better in those scenarios.
Finally, we add the second stage of training and train skip connections to obtain our final model. 
The color discrepancies disappear after this stage of training.
Interestingly, our not-diverse model - Model 1 achieves better scores than our diverse model - Model 4. 
However, when second-stage training is added, the diverse model achieves better results.
That is because the diverse model in the first stage has more difficulties outputting perfectly semantically consistent pixels but it outputs meaningful and diverse parts.
Especially when the erased parts are large, the diverse model does a better job as was discussed for Fig. \ref{fig:mask_plot} because it augments the encoded features with sampled ones. 
The color discrepancies are resolved with the second stage of training bringing the diverse model to a better score than the non-diverse model.

\textbf{Editing Results.}
For our editing results, we first encode the erased image and randomly sample a $W+$ to our mixing network. We edit the output of the mixing network with InterfaceGAN directions \cite{shen2020interpreting}.
Editing results are shown in Fig. \ref{fig:editing_results}.
We show smile, bangs, and beard addition results as well as image edits with blush eyebrows, blonde hair, and lipstick addition.
Our model can successfully both inpaint and edit erased images thanks to the semantically rich feature representations of the StyleGAN model bringing more capabilities under one framework.

\section{Conclusion}

In this work, we present a novel framework that can achieve diverse image inpainting and editing with GAN inversion.
To the best of our knowledge, our framework is the first one that can achieve those simultaneously. 
We present an extensive ablation study to show the effectiveness of our contributions. Our comparisons with competing methods show that 
we achieve significantly better scores than previous works both in terms of diversity and quality.

\section*{Acknowledgement}

This work has been funded by The Scientific and Technological Research Council of Turkey (TUBITAK), 3501 Research Project under Grant
No 121E097.



\appendix

\section{Architecture Details}
\label{sec:training}

\begin{figure}[]
    \centering
    \includegraphics[width=1\linewidth]{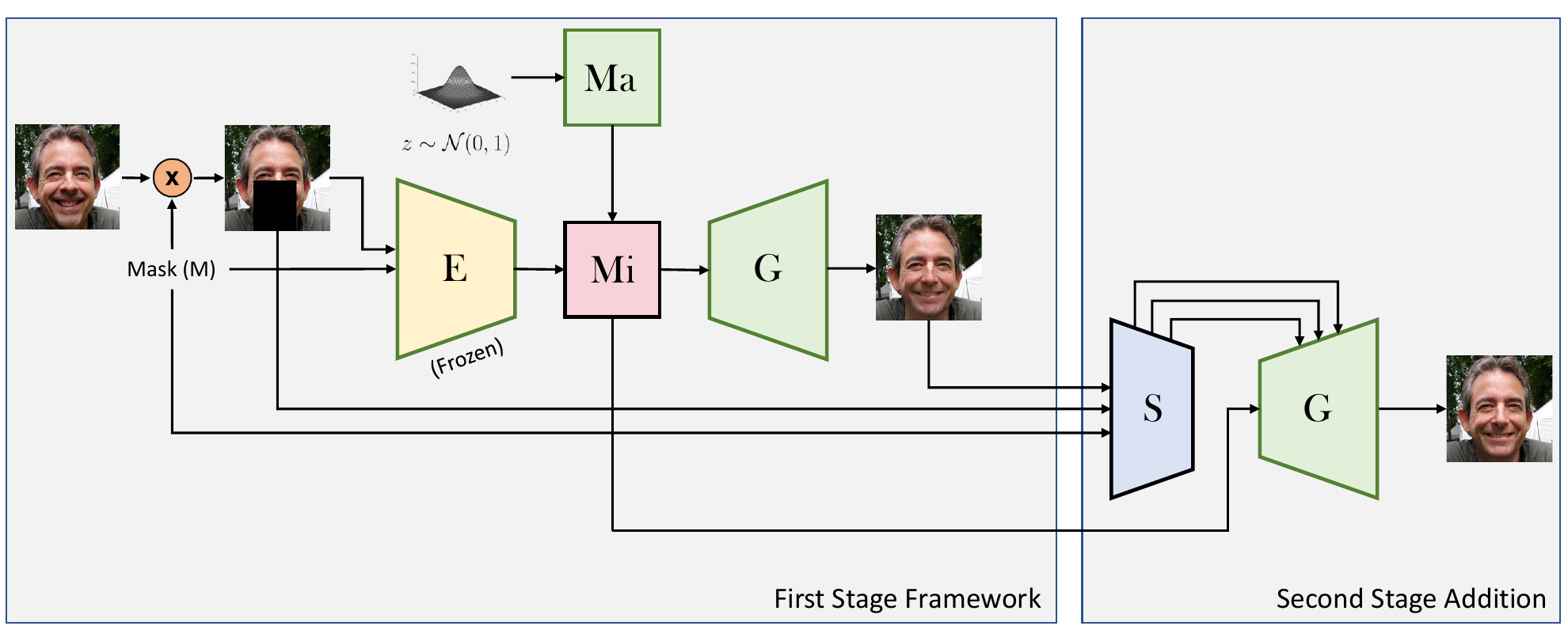}
    \caption{The overall architecture that is used in the second stage framework training. After learning the first stage model that includes $E$ and $Mi$, we learn skip connections from skip network $S$ to the generator $G$ to achieve high-fidelity reconstructions and seamless transitions across the boundaries of the masks.
    The same $\text{W}^{out}$ from the Mixing Network is used for both stages.}
    \label{fig:sup_overall}
\end{figure}

The final architecture is given in Fig. \ref{fig:sup_overall}. We follow a two-stage training pipeline. In the first stage, we train the Encoder and Mixing network. The architectures of them are as follows:

\textbf{Encoder ($E$).}
We adopt the encoder architecture from pSp \cite{richardson2021encoding} with minor modifications. First, we increase the first layer input channel number from $3$ to $4$ for taking the mask as an additional input. Then, we disable the normalization layers since we observe they decrease the performance of the model given that many input pixels may be $0$ due to removal of them.

\textbf{Mixing Network ($Mi$).}
We equip the mixing network with a neural network and gating mechanism.


The dimensions of $\text{W}^{enc}$ and $\text{W}^{rand}$ are both $14\times512$.
For the neural network $\text{NN}$, we use $14$ fully connected layers.
Each of them takes a style vector from $\text{W}^{enc}$ and $\text{W}^{rand}$ that are in $1\times512$ dimension. Fully connected layers have input and output dimensions of $512$.

In the second stage training, we set a new encoder which we refer to as Skip Encoder (S) in Fig. \ref{fig:sup_overall}. In the second stage training, we set the first Encoder frozen. We are interested in learning high-rate features and feed them to StyleGAN generator to achieve better fidelity to input image. 
The architecture is as follows: 

\textbf{Skip Encoder ($S$).}
The Skip Encoder takes input from the final output of the first stage model. We additionally feed the mask and erased input image to the Skip Encoder ($S$). 
They are concatenated and are fed to the $S$.
$S$ starts with a convolution layer to increase the channel size from $7$ to $32$ with a filter size of $3\times3$ and padding of $1$. The $32\times 256 \times 256$ feature maps are fed into residual blocks.
The residual blocks consist of three residual layers. Each residual layer consists of two convolution layers with batch normalization and parametric ReLu activation.
Each residual block downsamples the input resolution to half in its first residual layer using max pooling layer.
At each block, the channel size increases.
The Skip Encoder decreases the resolution to $32\times32$ at the end via 3 residual blocks.
The channel size at each block are as follows $48$, $64$, $96$ in the downsampling residual layers, respectively.
After we extract the $32$, $64$, and $128$ resolution feature maps, we pass them on $2$ more convolution blocks to retrieve skip connection addition ($G_{add}$) and multiplication ($G_{mult}$) maps, whose channels are compatible with the StyleGAN at respecting resolution. We do not have an activation function for $G_{add}$, but we have a sigmoid function for extracting $G_{mult}$. Lastly, StyleGAN generator features ($G_f$) are changed as follows:
\begin{equation}
\label{eq:skip}
    G_f = G_f + G_f * G_{mult} + G_{add}
\end{equation}

\section{Training Details.} We train the first stage for $500$k iterations with batch size of $8$ on two GPUs. We use learning rate of $1 \times 10^{-4}$ for all networks. We halve the learning rates at each $50$k iterations. We use the overall objectives given below to optimize the parameters of the Encoder ($E$) and the mixing network ($Mi$) as was also given in the main paper. 
We also use the same objective to optimize the parameters of $S$ in the second stage.

For both of the training stages, we use the same objective given in Eq. \ref{eqn:full_loss} and following hyperparameters, $\lambda_{a}=8\times10^{-2}$, $\lambda_{r1}=1$, and $\lambda_{r2}=1$.  
We set the pixel-wise and VGG reconstruction loss coefficients as $1$ and $5\times10^{-5}$, respectively for both ${L}_{rg}$ and ${L}_{gg}$.

{\small
\bibliographystyle{ieee_fullname}
\bibliography{egbib}
}


\end{document}